\documentclass{article}
\usepackage{graphicx} 
\usepackage{rotating}
\usepackage{url}
\usepackage{amsmath}%
\usepackage{MnSymbol}%
\usepackage{wasysym}%

\title{A Knowledge Engineering Primer\footnote{This primer can freely be used, shared and adapted under the terms of Creative Commons CC BY 4.0 license.}}
\author{Agnieszka Ławrynowicz\\
Poznan University of Technology, Poland}
\date{March 2024}

\begin{document}

\maketitle

\emph{The aim of this primer is to introduce the subject of knowledge engineering in a concise but synthetic way to develop the reader's intuition about the area.} 

\section{Introduction}
Knowledge can take different forms. 
We distinguish between declarative knowledge (knowing something) or procedural knowledge (knowing how, know-how), sensorimotor knowledge (riding a bicycle), and affective knowledge (deep understanding).
The classic definition of \emph{knowledge} derived from philosophy defines knowledge as a justified true belief. It can be said to occur in situations where we consider something to be objectively ``true'' or ``stated''. Another definition refers to what is ``explicit knowledge'' that is known and can be written down~\cite{NonakaTakeuchi1995aa}.

\emph{Knowledge representation}~\cite{BrachmanLevesque04} is a (symbolic) encoding of statements (or facts) that an agent (human or computer program) recognizes as true.
A mapping can be defined between the facts and their representation, assigning to the facts the corresponding symbols in the representation. Knowledge representation in artificial intelligence refers to how data, information, and knowledge are stored and processed in computer systems.

A \emph{knowledge base, $KB$} is, in some simplification, a collection of facts representing entities, classes, attributes, and relationships, relevant generally or in a particular domain, which is prepared in a digital form. 
The above definition of a knowledge base may resemble a database description. How, then, does a knowledge base differ from a database?
A good knowledge representation system, with which we represent a given knowledge base, on top of the ability to represent the required forms of knowledge, should ensure the power and efficiency of reasoning and knowledge extraction.
To perform inference, a type of software called a \emph{reasoner} is used to derive new facts from a set of already existing, explicitly represented facts or axioms.

It is worth noting the trade-off between the expressivity of the knowledge representation language (that is, the variety and number of possibilities for representing knowledge in it) and the performance of reasoning engines. The more complex the language, and thus the more diverse forms of modelled knowledge, the more complex the inference algorithms and the longer the inference time.
In \cite{kgcookbook}, Blumauer and Nagy classified popular knowledge organization systems. We extend this classification in   Table~\ref{tab:systemwiedzy} by additional forms of knowledge representation, showing selected classes of such systems of increasing complexity.

\begin{table}[!htb]
    \centering
        \caption{Knowledge organization systems.}
    \begin{tabular}{|p{2cm}|p{3.7cm}|p{5cm}|}
    \hline
& &    \\
        \textbf{Knowledge}

        \textbf{organization}

        \textbf{system} &
        \textbf{Building}

\textbf{blocks} &
\textbf{Examples} \\
\hline
        Thesaurus &
        Synonyms,

antonyms,

broader and narrower concepts,

associative relationships  &
blueberry = bilberry

cold $\neq$ warm

Blueberry juice \emph{is related to} blueberry
\\
\hline
 Taxonomy &    Hierarchical relationships &
 Blueberry \texttt{is-a} fruit

\\

\hline
Semantic network & Any unary and binary relations &
 Maria Skłodowska-Curie \texttt{was born in the year} 1867

 Parsley root \texttt{is part of} parsley
\\
\hline
Frame & Attributes inherited by subclasses and instances
&
Country \texttt{has a capital}.

Poland \texttt{has a capital} Warsaw
\\
\hline

Ontology & Classes, attributes, relationships, constraints &
Parsley \texttt{is a subclass of the class} plants

Carrot \texttt{has colour} orange

Blueberry juice \texttt{is made from} blueberry

\texttt{Every} tree \texttt{has at least one} root
\\
\hline
Knowledge graph & Classes, attributes, relationships, constraints, links to other knowledge bases &
Entity \texttt{Maria\_Sklodowska\_Curie} \texttt{is the same as} entity \texttt{wikidata:Q7186}
\\
\hline

    \end{tabular}

    \label{tab:systemwiedzy}
\end{table}

Issues of \emph{knowledge acquisition}, including issues of the knowledge base construction process, are dealt with by the field of \emph{knowledge engineering}.
A knowledge engineer explores a domain, determines which concepts are relevant to that domain, and creates a formal representation of entities, relationships, and constraints for that domain. Although he or she is often not a domain expert, the role of the knowledge engineer is to obtain knowledge from domain experts, among other sources.

Most knowledge representation systems proposed in artificial intelligence research are systems where knowledge is represented in symbolic form and easily readable by humans. The most important of these are:
\begin{itemize}
\item production rules~\cite{davis1977production},
\item semantic networks~\cite{richens1956preprogramming},
\item frames~\cite{10.5555/889222},
\item ontologies~\cite{Gruber1993ATA,Guarino1998},
\item knowledge graphs~\cite{DBLP:journals/csur/HoganBCdMGKGNNN21}. 
\end{itemize}

Most modern knowledge representation languages are declarative based on the concept of frames or first-order logic (predicate calculus)~\cite{newell1956logic,inbookK1,inbookK2}. Establishing a given language on the foundations of logic allows for the formalization and standardization of reasoning procedures, which in turn allows for constructing reasoning engines that operate on a given formalism.
It is also worth mentioning that while knowledge structures, such as knowledge graphs, are currently represented in symbolic form in order to operate on them, sub-symbolic representations are often created, such as \emph{embeddings}.

Why is the issue of representation important at all? What makes one representation better than another in the context of artificial intelligence? Many information and knowledge processing tasks can be very easy or complex, depending on how they are represented. This general principle applies both in everyday life and artificial intelligence. 
To illustrate it, let us take maps as an example. 
Old maps, such as those from the 16th century, were static, and their localities were marked as points rather than regions (polygons). 
The digital version of such maps, while continuing to mark localities as points, allows queries to be made to a historical-geographical information system about the location and attributes of localities over time and faster information retrieval.
Modern digital maps also have layers (buildings, roads, forests, etc.) and allow querying on various aspects of the terrain, such as Points of Interest etc. 
Depending on the representation and its expressivity, we can ask the model about different properties (e.g., only the geographical coordinates or also the type of Point of Interest some object may have).



We have already mentioned that a good knowledge representation system should facilitate both the acquisition of knowledge and its use, including reasoning based on its representation. In general, a good representation of both knowledge and information facilitates the subsequent task on which the representation is operated and increases the efficiency or speed of its solution, such as being easier to process by machine learning models or making it easier to answer questions. And it is often in terms of the task that we choose the suitable representation.

For example, assume that we want to build a machine learning model to recognize images of animals. In that case, a good data representation might be images in the form of raw pixels. The model will be able to learn from this form of input data and will be able to recognize different animals based on pixel patterns. If, on the other hand, we want to explore relationships between known scientists, a good option would be to create a graph in which the nodes are individuals, and the edges are labelled with the types of relationships between them (e.g., supervisor, co-author). Representing the data in this way makes finding connections between scientists and determining the degree of their proximity more simple. For example, we can easily find people who have published scientific articles together.

Other important aspects are the interpretability and reusability of a given knowledge model, including ease of modification and addition of new information. An example of a good knowledge representation in the medical domain could be a knowledge graph containing information on various diseases, symptoms and treatments. For example, a knowledge graph might contain information about various diseases, such as diabetes or heart disease, the typical symptoms of these diseases and what treatments are available. In this application, the knowledge graph makes it easy to find and interpret information about specific diseases and treatments and easily add new information.

An appropriate knowledge representation should facilitate reasoning. 
For example, in a health and nutrition ontology, one can define concepts such as:~\texttt{disease}, ~\texttt{diet}, ~\texttt{nutrient}, and ~\texttt{drug}, as well as relationships between them, such as: ~\texttt{diet supporting the treatment of the disease}, ~\texttt{drug for the disease} or ~\texttt{effect of the drug on the absorption of the nutrient}, and constraints, such as the ~state of~\texttt{hyperglycemia}, concerning blood sugar levels, is disjoint with the state of~\texttt{hypoglycemia}. This representation of knowledge in an information system, where individual concepts, relationships, and constraints (axioms) are explicitly represented, should facilitate reasoning about recommended diets for people struggling with a particular disease. A diet recommendation system, for example, could include as a component an ontology about diabetes and use it to infer that diabetes is a disease that requires a special diet and that certain nutrients are particularly important to be included in the diet for people with diabetes and certain others should be restricted. As a result, the system can generate consistent and reasonable dietary recommendations for such people.
Many knowledge representation systems are based on logic to facilitate inference and ensure that the generated conclusions or recommendations are consistent and verifiable.

\section{Logical foundations
}
\emph{Logics} are formal languages used to represent information so that conclusions can be drawn.
 \emph{Syntax} defines the forms of statements (sentences) that can be formulated in a given language (knowledge representation structures). \emph{Semantics} defines the \emph{meaning} of statements, their \emph{interpretation}, i.e., it determines the \emph{truth} of a statements in the world.
Statements in logical form represent certain aspects of the world. On the other hand, the world is the interpretation that gives meaning (semantics) to statements in logical form.
Meaning in the logical sense is the relationship between statements in logical form and interpretations, i.e. possible worlds, including imagined ones.

Logicians typically think in terms of \emph{model} theories~\cite{tarski1954}. Models are formally structured worlds against which truthfulness can be evaluated.
We say that $m$ is a model of the statement $\alpha$ if $\alpha$ is true in $m$.
By $M(\alpha)$ let us denote the set of all models of $\alpha$.
The \emph{entailment (logical consequence)} means that one thing follows (logically) from another: 
\begin{equation}
    KB \models \alpha
\end{equation}

The statement $\alpha$ is a logical consequence of the knowledge base $KB$ if and only if it is true in all worlds where $KB$ is true.
$KB \models \alpha$ if and only if $M(KB) \subseteq M(\alpha)$. A sentence is \emph{valid} if true in every model.

The definition of logical consequence can be applied to derive inferences,~i.e., performing \emph{logical inference}.
To understand the connection between the concept of logical consequence and logical inference, we can think of sentences that are logical consequences of the $KB$ knowledge base, of which there may be many. Inference algorithms find a subset of such statements as inferences (conclusions).
We say that an inference algorithm $i$ can \emph{derive} a statement $\alpha$ from the $KB$ knowledge base. 
An inference algorithm is \emph{sound} if only valid sentences are provable using it. An inference algorithm is \emph{complete} if every valid sentence is provable using it. 

As described above, the semantics of first-order logic is typically defined using model theory.
Statements can be assigned a logical value by defining the interpretation of the symbols of a given language belonging to the family of first-order logic.
Interpretation $\mathcal{I}$=($\Delta^\mathcal{I}, \cdotp^\mathcal{I}$) consists of a non-empty set  $\Delta^\mathcal{I}$ (domain of interpretation) and  interpretation function $\cdotp^\mathcal{I}$.
The interpretation function $\cdotp^\mathcal{I}$ assigns elements from the set of the symbols to $\Delta^\mathcal{I}$.

\section{Semantic networks} 
\begin{figure}[!h]
    \centering
    \includegraphics[width=0.8\textwidth]{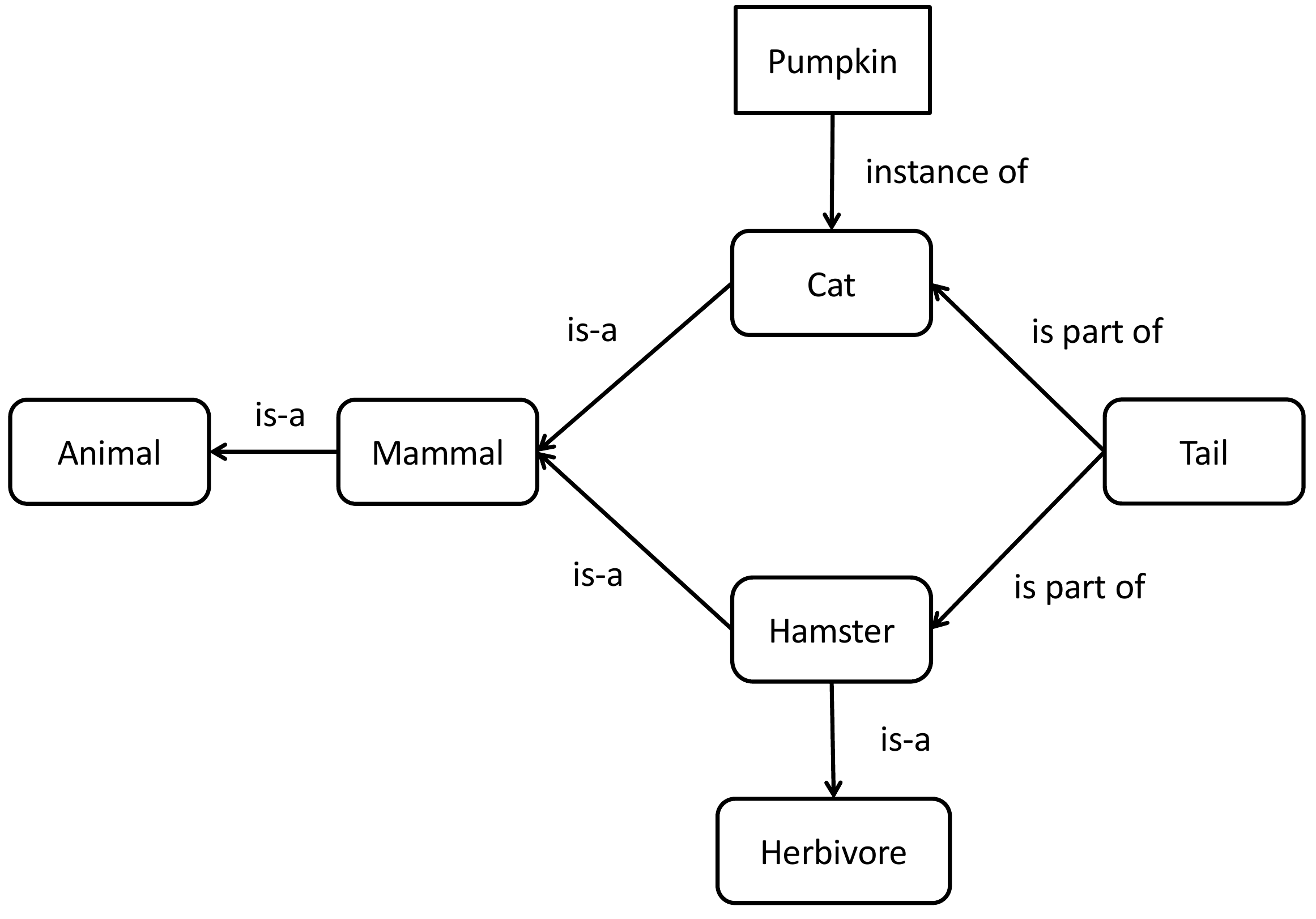}
    \caption{An example of a semantic network.}.
    \label{fig:semanticnet}
\end{figure}

\emph{Semantic networks} are a graphical notation for representing knowledge represented as a set of nodes (concepts) connected by labelled arcs that represent relationships between nodes. Figure~\ref{fig:semanticnet} shows an example of a semantic network.

Semantic networks became part of artificial intelligence research in the 1960s,  however, they had already been used in philosophy, psychology and linguistics.

The modern manifestation of semantic networks is~\emph{Semantic Web}~\cite{berners2001semantic}. The technologies of the Semantic Web, including the Resource Description Framework (RDF), RDF Schema (RDFS) and the Web Ontology Language (OWL), described in the following sections of this primer, are used to create and disseminate semantic networks and ontologies in a standardized form to allow global knowledge exchange on the World Wide Web. 
The father of the idea of the Semantic Web is Sir Tim Berners-Lee, who has also invented the World Wide Web.

\emph{Resource Description Framework, RDF} (\url{https://www.w3.org/TR/REC-rdf-syntax/}) combines the concept of semantic networks with web technologies. The resources we want to describe can be, for example, specific people, locations, or abstract concepts.

\begin{figure}[!htb]
    \centering
    \includegraphics[width=0.6\textwidth]{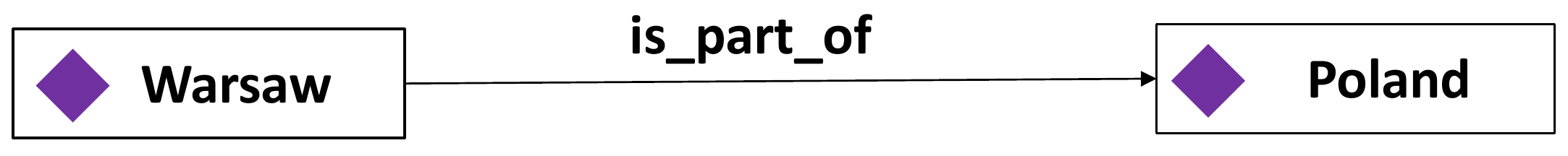}
    \caption{An RDF triple represented as a graph.}
    \label{fig:trojka_rdf}
\end{figure}

Data in the RDF model is represented using the so-called ''triple model'', where sentences are represented as triples consisting of a subject $s$, a predicate $p$ and an object $o$, for example:
\begin{center}
\texttt{Warsaw is\_part\_of Poland}
\end{center}
where \texttt{Warsaw} is the subject of the triple,
\texttt{is\_part\_of} is the predicate, and  \texttt{Poland} is the object.
Such a triple can be visualized as a graph~(Figure~\ref{fig:trojka_rdf}).\footnote{The graphical notation is inspired by the one used in a free, open-source ontology editor and framework Protégé (\url{https://protege.stanford.edu})}

Generally, there may be several localities with the same name, so the need arises to mark what resource we are referring to in a unique way.
Web technologies come to the rescue, in particular \emph{global identifiers (Uniform Resource Identifier, URI)}.
Using URIs, we can easily create globally unique names in a
decentralized manner -- each domain name owner can create new URI references.
URIs can also serve as a means of accessing
information describing a given resource, much like, known from web technologies, an URL (each URL is also a special case of a URI).
A URI may contain a part, called a fragment identifier, separated from the base part of the URI by the \# symbol.
For example, the role of the fragment identifier in the URI  \texttt{http://example.edu\#Warsaw} plays the string \texttt{Warsaw}.
Since URI identifiers are usually long character strings, a simplified, abbreviated version called qnames was introduced.
An URI expressed as a qname consists of two parts: the namespace and the identifier, separated by a colon. For instance, in \texttt{edu: Warsaw}, a qname identifier, which refers to the namespace, is \texttt{edu}, while \texttt{Warsaw} refers to the fragment identifier.

The basic elements of RDF are:
\begin{itemize}
    \item resources, which are identified by URIs and correspond to nodes in the graph, e.g.~\texttt{http://example.edu},
    \item blank nodes, i.e.~graph elements that are not given a label or URI identifier, and are often used to describe objects that do not have their own URI identifier or to build complex expressions that consist of multiple elements, when at the same time one does not want to create separate resources for each element. The strings representing blank nodes start with the characters \texttt{\_:}, and software frameworks create them automatically,
    \item properties, identified by URIs, corresponding to arcs in the graph, and representing the binary relationships, e.g.~ \texttt{http://example.edu\#is\_part\_of},
    \item literals that represent specific data values e.g. \texttt{Warsaw}, \texttt{2022-05-26}.
\end{itemize}

Now let us formalize our knowledge of RDF graphs. Let us consider the pairwise disjoint sets $\mathbf{U}$, $\mathbf{B}$ and $\mathbf{L}$.
They denote resources (URI references), blank nodes, and literals.
\emph{RDF triple} is a tuple $t = (s, p, o) \in (\mathbf{U} \cup \mathbf{B}) \times \mathbf{U} \times (\mathbf{U} \cup \mathbf{B} \cup \mathbf{L})$, where $s$ is the subject, $p$ is the predicate, and $o$ is the object of the triple.
\emph{RDF graph} (or RDF dataset) $\mathcal{G}$ is a set of RDF triples.

Since we only deal with, at most, binary relations in an RDF graph, how can we represent relations that are inherently $n$-ary in such a graph? We can use the \emph{reification} design pattern, illustrated in the following example.

\textbf{Example 1 (Reification).}
Now, let us examine the task of transforming a given table from a relational database into an RDF graph that reflects the meaning and relationships of the data in the table.
For the purposes of our task, we will use Table~\ref{tab:rdf_nray}, which contains shopping data.

\begin{table}[!htb]
    \caption{Data on purchases.}
    \centering
    \begin{tabular}{|l | l | l | l|}
    \hline
    \textbf{Buyer} & \textbf{Seller} & \textbf{Product} & \textbf{Number of pieces} \\
\hline
    Marcin Kowalski & Shop1 & Natural yoghurt & 5 \\
    Aleksandra Nowak & Shop2 & Butter & 2 \\
\hline
    \end{tabular}
    \label{tab:rdf_nray}
\end{table}

In the case under consideration, the 'purchase' relationship has more than one participant, being in this relationship.
In addition, none of the table's columns stands out as leading for the relationship (purchase).
When we encounter such a situation, it is worth using the following design pattern, the so-called \emph{reification}.
We create an instance representing a relation with links to all instances that are in this relation.
We will represent the $n$-ary relation in the RDF model by creating just such a new instance representing the relation between $n$ specific instances.
\begin{figure} [!htb]
    \centering
    \includegraphics[width=1.0\textwidth]{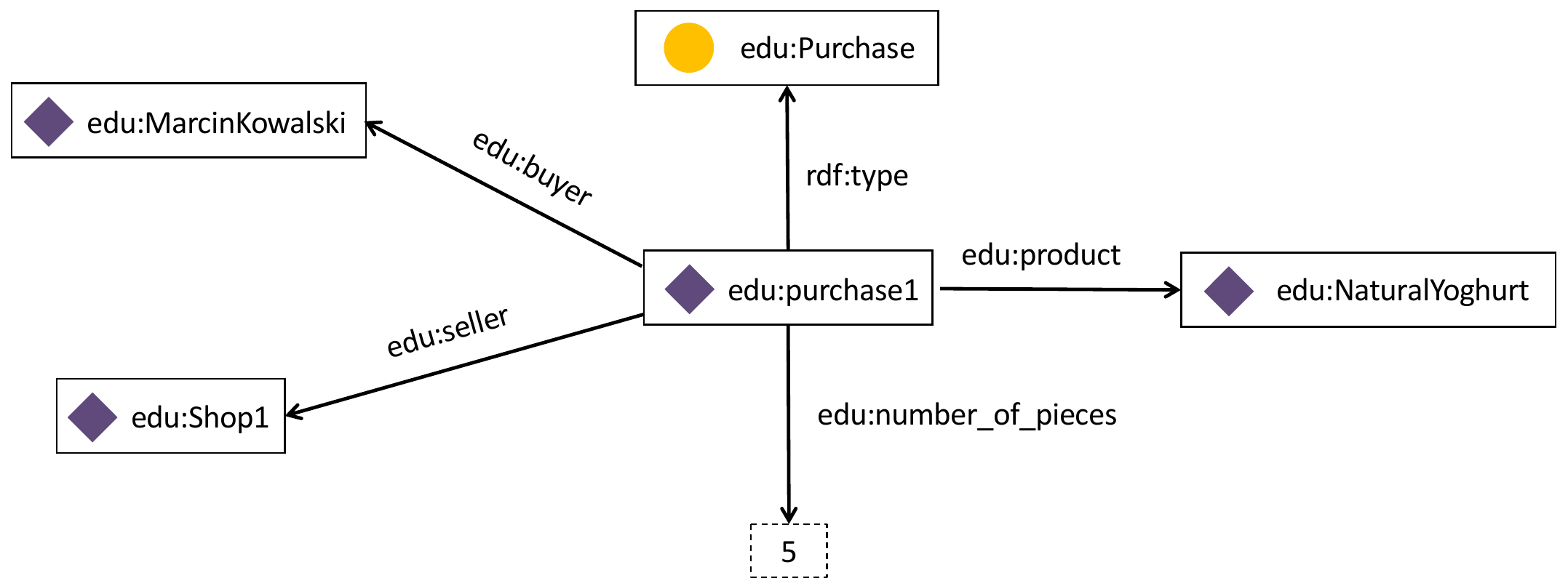}
    \caption{RDF graph depicting $n$-ary purchase relationship.}
    \label{fig:rdf_n_ary}
\end{figure}

This requires creating a total of $n$+1 triples: one to create the main instance of the relationship and one for each object that is in that relationship.
Following the given pattern, and assuming that our namespace will be \texttt{http://example.edu}, the first row of the table can be visualized as an RDF graph as in the Figure~\ref{fig:rdf_n_ary}. The set of triples corresponding to this graph is as follows:
\begin{verbatim}
edu:purchase1 rdf:type edu:Purchase
edu:purchase1 edu:product edu:NaturalYoghurt
edu:purchase1 edu:number_of_pieces "5"
edu:purchase1 edu:buyer edu:MarcinKowalski
edu:purchase1 edu:seller edu:Shop1
\end{verbatim}
$\blacksquare$

\section{Frames}

\emph{Frame} is a complex data structure used in artificial intelligence to represent stereotypical situations or events.
Frames, as a form of knowledge representation, are derived from semantic networks and are a specific version of them.
A frame is a representation of an object or category and allows one to collect information about them. It has attributes and is in relationships with other objects or categories.
This way of representing and organizing knowledge reflects the structure of the real world. 
Two types of frames can be distinguished:
\begin{itemize}
 \item individual (represent a single object, such as a specific city),
 \item general (represent a category of objects, e.g., cities).
\end{itemize}

A single frame is a named list of \emph{slots} that are filled with \emph{facets}. Slots are individual pieces of information that make up a given frame, such as:
\begin{verbatim}
   (frame-name
       <slot-name1 facet1>
       <slot-name2 facet2> ...)
\end{verbatim}

A frame reflects the previously accumulated experience of specific situations through defined and default values.
General frames have a slot~\texttt{IS-A}, which is filled in with the name of another general frame, such as:
\begin{verbatim}
(Carrots
  <:IS-A Vegetables>
  <:colour orange>  ...)
\end{verbatim}

More specific frames inherit facets from more general frames.
Individual frames have a slot~\texttt{INSTANCE-OF}, which is filled with the name of the general frame, e.g.:
\begin{verbatim}
(warsaw
  <:INSTANCE-OF City>
  <:voivodeship mazowieckie>
  <:population 1 860 281>  ...)
\end{verbatim}

Inference using a frame is done by:
\begin{itemize}
\item consistency checking when filling a slot with a value,
\item inheritance of defined and default values (according to \texttt{IS-A}, \texttt{INSTANCE-OF}).
\end{itemize}

One way frame-based knowledge representation manifests itself in modern systems is through the use of frame semantics. Frame semantics is an approach to natural language processing that uses frame-based representations to understand or generate natural language. It can be used in applications such as information extraction, machine translation, and dialogue systems.
An example of the practical use of this idea is FrameNet~\cite{10.3115/980845.980860}, a lexical database of English containing syntactically and semantically labelled examples of sentences from a corpus of texts.
FrameNet consists of so-called semantic frames.
A semantic frame is a description of the type of event, relation or entity and the units that constitute it.
It consists of frame elements (frame roles) and lexical units, i.e., words that are found in the text invoke the frame.
In addition, sentences from the corpus annotated with frame elements are attached to the frame.

\textbf{Example 2.}
Consider an example of a frame called~\texttt{Cooking}, which represents the general concept of cooking and contains a number of slots that represent different aspects of cooking, such as a cook, ingredients, cooking method and cooking equipment. Some examples of lexical units (words or phrases that can invoke the frame) include \texttt{cooking}, \texttt{baking}, \texttt{grilling}, etc.
Some examples of frame elements, specific slots in the frame that represent different aspects of a concept, are~\texttt{Cook}, \texttt{Produced\_food}, \texttt{Ingredients}, \texttt{Container}. 

A sample sentence that fits into this example, where we can find the lexical unit associated with the frame \texttt{Cooking} might look like this:
\\
\texttt{Today Maria is going to (bake)} [the lexical unit that invokes the frame] a \texttt{raspberry cake.}
\\
In this sentence, we have an example of a lexical unit \texttt{bake} that triggers the frame \texttt{Cooking} as well as frame elements such as \texttt{Produced\_food}  (\texttt{raspberry cake}) and \texttt{Cook} (\texttt{Maria}). Some frame elements (e.g., \texttt{Container}) are not specified. \\
$\blacksquare$

In a dialogue system, frame semantics can be used to help the system understand the user's intent. For example, if a user asks a question about the weather, the system can use frame semantics to identify the appropriate frame associated with the intent (e.g., ~\texttt{weather\_forecast}) and fill in the appropriate fields associated with that intent (e.g., location, date) to generate a response.

Frame semantics is also used in Wikipedia, where it represents relationships between different concepts and provides context for articles. For example, a Wikipedia article about a particular city might include a frame (called an Infobox) that contains information about the city's location, population, and other typical data. This allows readers to more easily access additional related information.

Knowledge representation in the form of frames and frame semantics is also used in robotics to help robots understand their environment and interact with it. In this context, frames represent various concepts and objects a robot may encounter, such as furniture or tools. In addition, the slots in each frame can contain information about the object's physical characteristics, such as size, shape and colour, as well as its function and use. Frame-based inference systems may allow robots to use knowledge about the environment to make decisions and take actions. For example, a robot can use its knowledge of objects in a room to navigate through it or to recognize and identify specific objects.
Frame semantics can also be used in natural language processing, allowing robots to understand and respond to human commands and questions. For example, a robot can use frame semantics to understand the intent behind the command ``pick up the red cup'' by identifying the appropriate frame (e.g., \texttt{object\_manipulation}) and filling in the appropriate fields (e.g., object to be manipulated, object's features).

Knowledge representation using frames played a significant role in developing early ontologies and tools for knowledge representation and manipulation. One example is the Protégé system, a popular software platform for building, editing and manipulating ontologies.
In Protégé, a knowledge representation framework is used to represent concepts and their relationships in the form of frames, which consist of a collection of slots and fillers. Each frame represents a specific concept, and the slots represent properties or characteristics of that concept. Fillers are specific values that are assigned to each slot.

Using frame knowledge representation and frame semantics in ontologies and tools such as Protégé helped provide a structured and organized way to represent and manipulate knowledge, enabling users to manage and use large amounts of information more effectively.

\section{Ontologies}

The word \emph{ontology} originated in philosophy. Ontology, as a philosophical discipline, was first studied by Aristotle. In his important philosophical work ``Metaphysics'' Aristotle defined what later became known as ontology, or the science of ``being as being'', which studies the nature of entities and their attributes. Ontology deals with describing and categorizing entities based on their structure and properties.

In computer science, ontologies are formal representations of concepts, relationships, and constraints within a domain that facilitate communication and understanding between multiple parties.
In computer science, ontology is an engineering artefact, i.e.,~not the study of entities and their categorization but the concrete result of such categorization. Therefore, although from a philosophical perspective, ontology can be seen as a particular system of categories responsible for a certain vision of the world, independent of its representation, in computer science, an ontology is dependent on the languages used to represent it.
Several definitions of ontology have been proposed in this context. Ontology is defined by Gruber as a ``formal, explicit specification of a shared conceptualization of a domain of interest''~\cite{Gruber1993ATA}.
Formality is about making the ontology explicit for interpretation by machines. Clarity refers to making sure that all concepts and their interrelationships are clearly defined.
Sharing refers to the ontology's capture of some consensus in modelling concepts, accepted by a community or several stakeholders, rather than an individual view. The domain of interest is established between the requirements of a specific application and the  ``unique truth''.
Another definition of ontology in computer science by Uschold states that it is ``the representation, formalization and specification of important concepts and relationships within a given domain''~\cite{Uschold1996}. This definition emphasizes the role of ontology in representing and specifying key concepts and relationships within a given domain, as well as formalizing these concepts and relationships.

Among the available ontologies, we can distinguish between foundational ontologies, i.e. ontologies that define the basic concepts and distinctions, the types of entities existing in the world and the relationships between them, such as objects fixed in time versus events or real or abstract objects. For example, the DOLCE~\cite{gangemi2002dolce} or BFO~\cite{10.5555/2846229} ontology fall into this category.

On the other hand, domain-specific ontologies are designed to represent domain-specific concepts and relationships. Examples of such ontologies include biomedical ontologies, such as the SNOMED ontology, which is used to represent medical concepts and their relationships, the GO ontology, which is used to represent gene functions, or CHEBI, an ontology of biologically relevant chemical compounds.

To model ontologies, a number of representation languages have been proposed that provide various kinds of features. The most widespread have been languages based on the first-order logic since it is equipped with formal semantics and thus facilitates machines' interpretability of encoded knowledge.

\subsection{RDFS language}
\begin{figure}[!htb]
    \centering
    \includegraphics[width=0.8\textwidth]{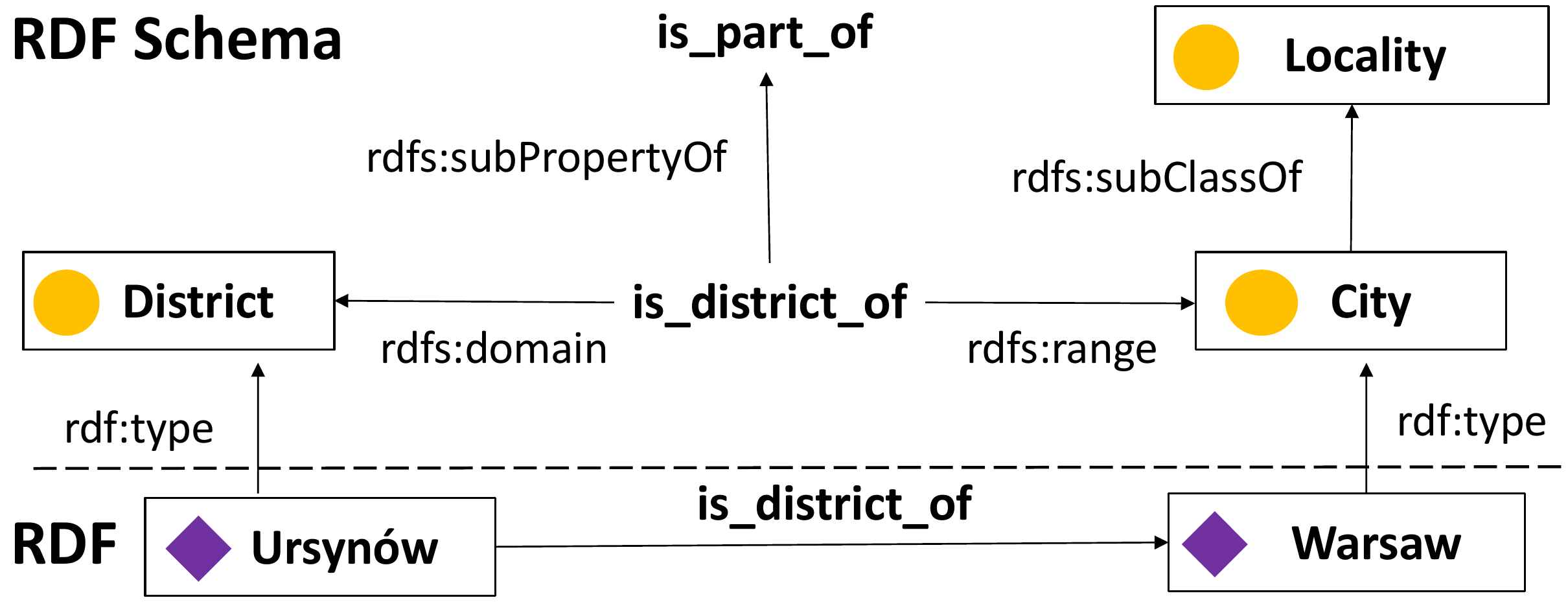}
    \caption{RDFS-represented data schema layer embedded on RDF-represented data layer, together forming a single semantic network.}.
    \label{fig:rdfrdfs}
\end{figure}

Simple ontologies can be represented using \emph{RDFS language (RDF Schema)}~(\url{https://www.w3.org/TR/rdf-schema/}).
RDFS belongs to the Semantic Web technology stack. It integrates with RDF by enriching data with its semantics formulated in the form of a data schema.
Figure~\ref{fig:rdfrdfs} illustrates an example of the semantic network, where RDF is the data layer and RDFS is the data schema layer.

RDFS introduces a distinctive vocabulary to indicate to inference (reasoning) engines what conclusions they should derive from the facts being modelled.
To model class hierarchies, RDFS includes the keyword \texttt{rdfs:subClassOf}, with which we can model that class $C_1$ is a subclass of class $C_2$. So, for example, we can model that class \texttt{City} is a subclass of class \texttt{Locality}. Continuing with the example, if then the knowledge base contains the following triple:
\begin{center}
\texttt{Warsaw rdf:type City}
\end{center}
then we can infer (deduce) that
\begin{center}
\texttt{Warsaw rdf:type Locality}
\end{center}
which allows us to model and infer the class hierarchy.

Similarly, we can use the property \texttt{rdfs:subPropertyOf} to model the fact that property $P_1$ is a subproperty of property $P_2$. For example, we can model that \texttt{is\_district\_of} is a subproperty of \texttt{is\_part\_of} and then given facts:
\begin{center}
\texttt{Ursynów is\_district\_of Warsaw}
\end{center}
then we can infer that
\begin{center}
\texttt{Ursynów is\_part\_of Warsaw}.
\end{center}

In addition, RDFS introduces a vocabulary for defining domain constraints (\texttt{rdfs:domain}) and range constraints (\texttt{rdfs:range}).
When we denote the domain of property $P_1$ as class $C_1$ and there is a fact 
\texttt{entity1} $P_1$ \texttt{entity2} 
in the knowledge base, the inference will be that the instance \texttt{entity1} belongs to class $C_1$.
Similarly, we can introduce a range constraint (\texttt{rdfs:range}) to limit the membership of a given class of objects (the third element) of the triple.
For example, if there are the following facts in the knowledge base:
\begin{center}
\texttt{is\_district\_of rdfs:range City}\\
\texttt{Ursynów is\_district\_of Warsaw}
\end{center}
then we can infer that
\begin{center}
\texttt{Warsaw rdf:type City}
\end{center}

\subsection{OWL ontology modeling language}
The most popular, standard ontology modeling language is the \emph{OWL (Web Ontology Language)}~(\url{https://www.w3.org/TR/owl-features/}).
Standardizing the knowledge representation language helps in creating tools for editing knowledge represented in the language and also in developing reasoning engines.
OWL allows describing concepts (classes) in a formal, unambiguous way, based on set theory and logic.
OWL ontologies are implementations of \emph{description logic}~\cite{DBLP:conf/dlog/2003handbook}, which is a subset of the first-order logic.

Knowledge bases represented using description logic typically consist of a terminological part, i.e. the schema of the knowledge base, and an assertional part, i.e. the data in the knowledge base.
\emph{The terminological part (terminological box, TBox)} contains the vocabulary used to describe the hierarchy of classes and relationships in the knowledge base.
\emph{The assertional part (assertional box, ABox)} contains statements about properties of instances.

OWL extends RDF and RDFS by providing additional vocabulary.
OWL can be written using RDF syntax, where expressions containing OWL vocabulary are embedded in RDF documents and interpreted according to OWL semantics.
Other common ways of writing OWL are turtle and the Manchester syntax.
References to syntax formalized with description logic can also often be encountered.

The main elements that make up an ontology represented in OWL are:
\begin{itemize}
\item \emph{entities} -- classes, properties, individuals (instances) and any other elements of the modelled domain. A class is interpreted as a set, a property as a binary relation, and an individual as an element of a set;
\item \emph{expressions} -- complex classes occurring in the modelled domain;
\item \emph{axioms} -- assertions that are true in the modelled domain.
\end{itemize}
The ontology represented in OWL is a set of axioms.

Classes represent collections of individuals (instances).
For example, by writing \texttt{City rdf:type owl:Class}, we can express that \texttt{City} is the class of cities, which includes instances such as \texttt{Warsaw}, \texttt{Poznan}, etc.
The two special classes are: the universal class~\texttt{owl:Thing}, which represents the set of all instances, is a superclass of all classes, and its interpretation is the entire domain under consideration, and the bottom class \texttt{owl:Nothing}, which represents the empty set of instances, i.e.~constraints that are impossible to satisfy all together (its interpretation is the empty set).

Complex classes can be built from simpler classes using logical operators. This gives us a ``conceptual Lego'', where we construct complex classes from other (potentially complex) classes. 

Let us denote (complex) classes by $C, D$, properties by $R, S$, and individuals (instances) by $a, b$. Examples of operators and their representation in description logic and turtle notations are shown in Table~\ref{tab:classconstructors}.

\noindent
\begin{sidewaystable}
\caption{Examples of logical operators that can be used in OWL to construct complex classes shown using description logic and turtle syntaxes.}\label{tab:classconstructors}

\centering
\begin{tabular}{l l p{6cm} p{7cm}}
\hline
Operator & Syntax  & Example  & Example     \\
 & (description logic)  & (description logic) & (turtle)    \\
\hline
	universal class	& $\top$ &  $\top$ & \texttt{owl:Thing}\\
	bottom class	& $\bot$ &  $\bot$ & \texttt{owl:Nothing}\\
	negation & $(\neg C)$ &   ( $\neg$ \texttt{Meat} ) & \texttt{owl:complementOf Meat} \\
	intersection &  $(C \sqcap D)$ &  ( \texttt{Child} $\sqcap$ \texttt{Man}) & \texttt{owl:intersectionOf (Child Man)}  \\
    union	& $(C \sqcup D)$ &  ( \texttt{Herbivore} $\sqcup$ \texttt{Carnivore} $\sqcup$~\texttt{Omnivore} ) & \texttt{owl:unionOf (Herbivore Carnivore Omnivore)}   \\
	existential & $(\exists R.C)$ & ( $\exists$\texttt{eats.Meat} ) & \texttt{owl:onProperty :eats ;}    \\
	quantification &   &   & \texttt{owl:someValuesFrom :Meat} \\
    value 	& $(\forall R.C)$ & $(\forall$\texttt{eats.VegetarianProduct} ) &\texttt{owl:onProperty :eats ;} \\
    restriction	&  & & \texttt{owl:allValuesFrom :VegetarianProduct}\\
\hline
\end{tabular}
\end{sidewaystable}

\begin{figure}[!htb]
    \centering
    \includegraphics[width=0.49\textwidth]{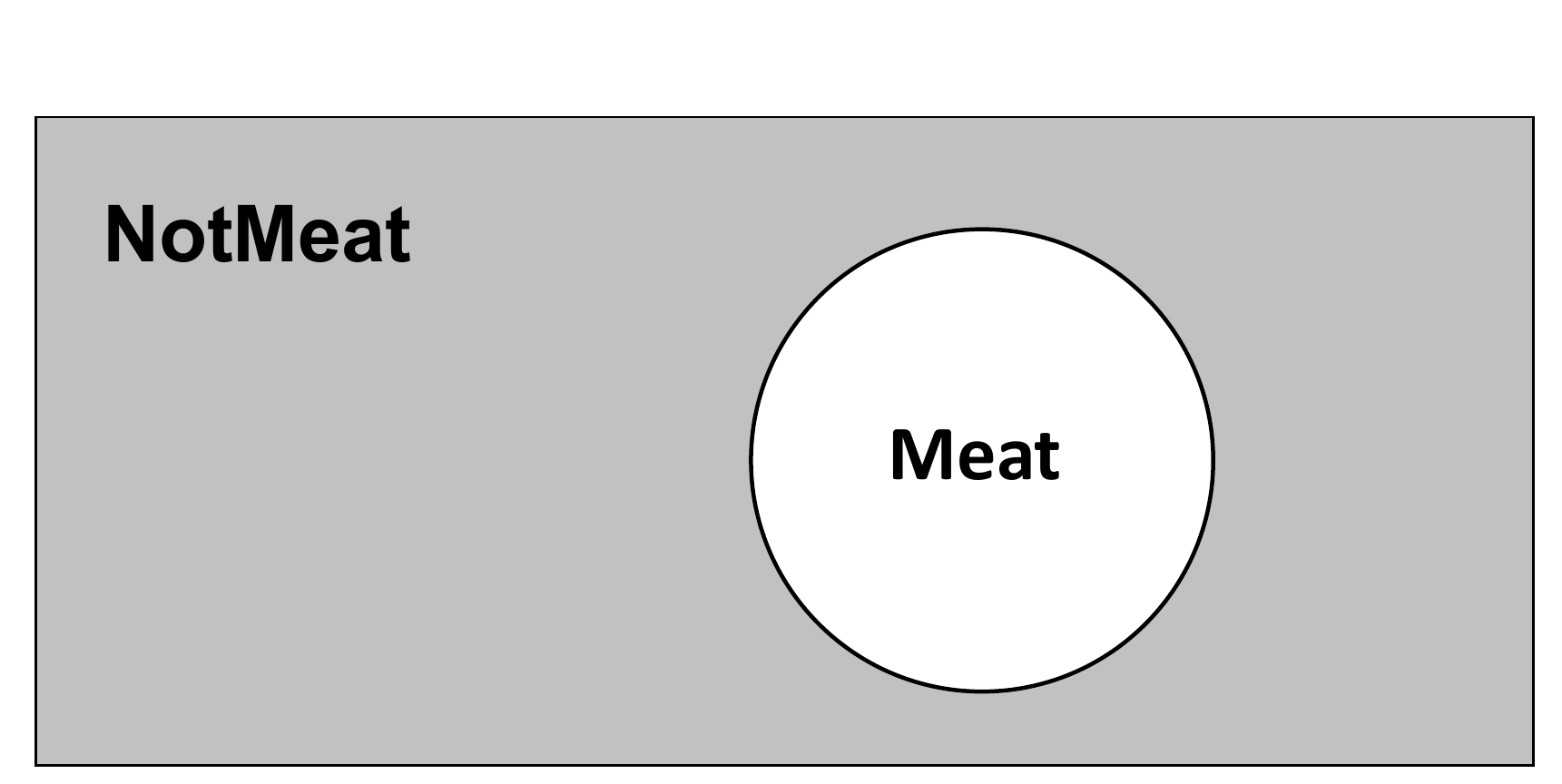}
    \includegraphics[width=0.49\textwidth]{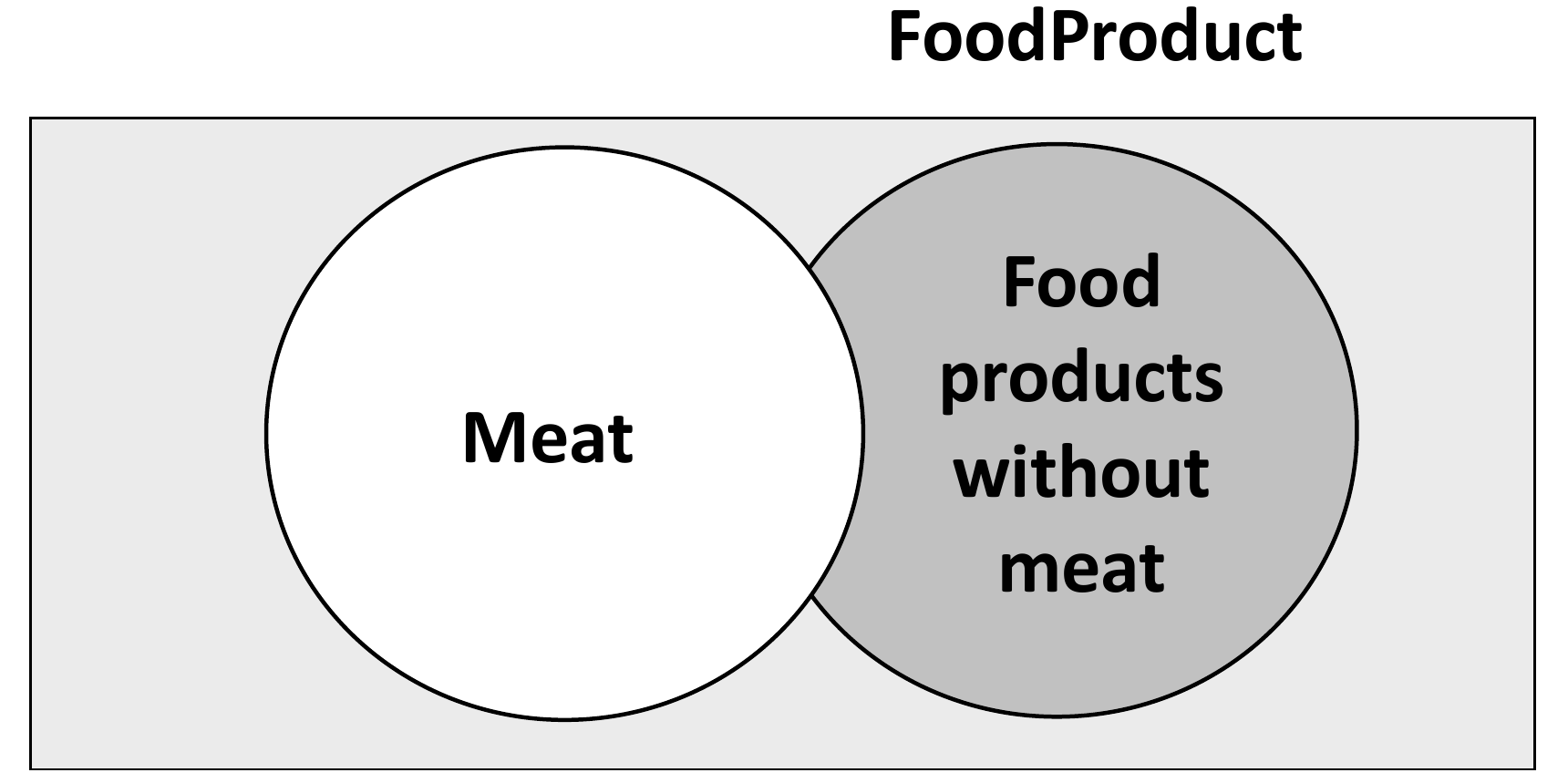}

    \caption{Interpretation of negation as complement (left) and~complement in the context of superclass (right).}
    \label{fig:nonmeat}
\end{figure}

Because of the \emph{Open World Assumption, OWA}, discussed later in this primer, it is worth noting the interpretation of some operators, particularly negation. Negation is interpreted in OWL as a complement, as illustrated in the Figure~\ref{fig:nonmeat} on the left, referring to the expression \texttt{owl:complementOf Meat}.

In practice, inferring all imaginable non-meat objects, including those unknown and not described in the knowledge base is challenging.
Therefore, to infer negation, it is often captured in the context of a superclass (e.g., \texttt{FoodProduct}), which describes a set of instances, some of which are, for example, meat and the rest are other instances belonging to the class \texttt{FoodProduct} as illustrated in the Figure~\ref{fig:nonmeat} (right), where the area marked in light gray is the complement of a white and dark gray circle relative to the rectangle surrounding the whole.
The class of things that vegetarians eat is in the area marked in dark gray.

Properties exist independently of classes.
Properties in OWL are divided into:
\begin{itemize}
\item object properties that link a resource to another resource, e.g. the property \texttt{is\_part\_of} can link instances \texttt{Ursynów} and \texttt{Warsaw},
\item data properties that connect a resource to a literal, e.g. the \texttt{weight} property can associate a given product with its weight,
\item annotation properties that link the resource to a note about it, e.g. \texttt{rdfs:label} links the resource to its label.
\end{itemize}

In addition, properties can have their own characteristics, of which we can distinguish:
\begin{itemize}
\item inverse property, e.g.:
\begin{verbatim}
:is_parent_of rdf:type owl:ObjectProperty ;
               owl:inverseOf :has_parent .
\end{verbatim}
\item functional property, e.g:
\begin{verbatim}
:has_father rdf:type owl:ObjectProperty ,
                 owl:FunctionalProperty .
\end{verbatim}
\item transitive property, e.g:
\begin{verbatim}
:is_part_of rdf:type owl:ObjectProperty ,
                      owl:TransitiveProperty ;
             rdfs:domain :Region ;
             rdfs:range :Region .
\end{verbatim}
\end{itemize}

For classes, we can define three main types of axioms that define relationships between classes (also shown in Table~\ref{tab:owlclassaxioms}):
\begin{itemize}
\item \emph{subsumption},
\item \emph{equivalence},
\item \emph{disjointness}.
\end{itemize}

\begin{table}[!htb]
\caption{The main axioms in the OWL language that define the relationships between classes presented by means of two syntaxes: the description logic syntax and the turtle syntax.}\label{tab:owlclassaxioms}
\centering\scriptsize%
\begin{tabular}{p{1.5cm} p{1.5cm} p{3.5cm} p{4cm}}
\hline
Axiom & Syntax  & Example  & Example     \\
             & (description   & (description & (turtle)    \\
             & logic) & logic) &    \\
\hline
subsumption	& $C \sqsubseteq D$ &	\texttt{Boy} $\mathsf{\sqsubseteq}$ \texttt{Child} &  \texttt{Boy owl:subClassOf Child} \\
equivalence  & $C \equiv D$ & \texttt{Country} $\mathsf{\equiv}$ \texttt{State} & \texttt{Country owl:equivalentClass State}\\
disjointness  & $\mathsf{C} \sqcap \mathsf{D} \sqsubseteq \bot$ & \texttt{Herbivore} $\sqcap$ \texttt{Carnivore} $\sqsubseteq \bot$  & \texttt{Herbivore owl:disjointWith Carnivore}
\\
\hline
\end{tabular}
\end{table}

Analogous axioms can be defined for properties, but they are much less frequently specified due to the \emph{entity-centric} characteristics of the ontologies represented in OWL.

We introduce the subsumption relation using the keyword \texttt{owl:subClassOf}. By modelling the subsumption relation, we introduce the necessary conditions that instances of a class must satisfy.
In this way, we model semantic class descriptions.
One can interpret such a relation as a one-way implication. For example, we may wish to model that ``all carnivores eat meat'':
\begin{verbatim}
:Carnivore rdf:type owl:Class ;
       rdfs:subClassOf [ rdf:type owl:Restriction ;
                        owl:onProperty :eats ;
                        owl:someValuesFrom :Meat
			   ] .
\end{verbatim}

The keyword \texttt{owl:equivalentClass} represents an equivalence relation.
We model class definitions in this way.
Such a relation can be interpreted as a two-way implication, which defines the necessary and sufficient conditions to consider an instance of a class as an instance of another class and vice versa. This means that equivalent classes share the same set of instances.
For example, we may want to model that ``every boy is a child and a man'' and at the same time ``everyone who is a child and a man is a boy'':
\begin{verbatim}
:Boy rdf:type owl:Class ;
	     owl:equivalentClass [ rdf:type owl:Class ;
			                   owl:intersectionOf ( :Child  :Man )
			 ] .
\end{verbatim}
As long as we do not explicitly introduce the \emph{disjointness constraint}, classes can share  instances.
If we want that a given instance cannot belong to two given classes at the same time, we can introduce the axiom of class disjointness using the keyword \texttt{owl:disjointWith}.
For example, given the following statements:
\begin{center}
\texttt{Herbivore owl:disjointWith Carnivore}
\\
\texttt{Pumpkin rdf:type Carnivore}
\end{center}
we can infer that Pumpkin is not a herbivore.
The introduction of disjointness constraints into ontologies is very important from the point of view of checking the consistency of an ontology or knowledge base.

At the level of individuals (i.e. instances of classes), there are two main types of axioms:
\begin{itemize}
\item assertions of individuals to classes, e.g.:~\texttt{Warsaw rdf:type City}, and
\item assertions of individuals to properties, e.g.:~\texttt{Ursynów is\_district\_of Warsaw}.
\end{itemize}

The \texttt{owl:sameAs} property has an important role in reconciling entities that have different identifiers but are semantically equivalent to each other.
In contrast, when wishing to express that given instance identifiers refer to different objects, we can use the property \texttt{owl:differentFrom} (modelling the relationship between two instances) or \texttt{owl:allDifferent} (modelling the relationship between instances from a set as pairwise disjoint).

\subsubsection{Reasoning}
The use of logic to model ontologies allows the use of inference engines.
We can use inference engines, e.g. to check that all statements and definitions in the ontology are mutually consistent, to check which classes are in a superclass-subclass relationship (subsumption relationship) with each other automatically, and others. Inference can thus keep the hierarchy of classes in the ontology in the right order.

The basic inference tasks in OWL can be divided into schema (,,terminological'' part of the ontology) inference tasks and instance (,,assertional'' part of the ontology) inference tasks.

For the terminology (i.e. TBox) these are:
\begin{itemize}
\item checking whether a given (complex) class $C$ is a subclass of another class $D$ in a logical sense (subsumption test), i.e. whether $C$ \texttt{owl:SubClassOf} $D$ is a logical consequence of $KB$ (the set of instances of class $C$ is a subset of instances of class $D$ in all $KB$ models),
\item checking whether a given (complex) class $C$ is logically equivalent to another class $D$ i.e. whether $C$ \texttt{owl:EquivalentClass} $D$ is a logical consequence of $KB$ (the set of instances of class $C$ is equal to the set of instances of class $D$ in all $KB$ models),
\item checking whether a class is \emph{satisfiable}, whether the set of constraints describing it is not contradictory, i.e. whether $C$ \texttt{owl:EquivalentClass owl:Nothing} is not a logical consequence of $KB$ (the set of instances of class $C$ is not an empty set for some $KB$ model).
\end{itemize}

The inference tasks typical of the assertional part, ABox, are:
\begin{itemize}
\item checking whether the ABox is \emph{consistent}, i.e. whether it has a model,
the so-called task of \emph{consistency checking},
\item checking whether a given individual $a$ is an instance of a given class $C$ in the logical sense, the so-called task of \emph{instance checking},
\item given an ABox and a class $C$, finding all individuals $a$ such that the assertion $a$ \texttt{rdf:type} of $C$ is a logical consequence of the ABox, so-called~\emph{retrieval problem},
\item having an individual $a$ and a set of classes, finding \emph{most specific class} $C$ from the set such that the assertion $a$ \texttt{rdf:type} $C$ is a logical consequence of ABox, so called \emph{realization problem}.
\end{itemize}

Satisfiability and consistency tests can be used to determine the meaningfulness of the $KB$ knowledge base.
Subsumption tests are often used to automatically construct class hierarchies.
Instance tests are queries designed to return individuals that satisfy the query conditions.

\subsubsection{``Closed world assumption'' versus ``open world assumption''}
Inference engines in OWL operate on certain assumptions, which sometimes differ from those made to query relational databases, for example.
When dealing with a standard, centralised database, the so-called~``closed-world assumption'' is made, i.e. that we have complete knowledge of the instances and that missing information is negative information (negation-as-failure).
However, by querying the knowledge base represented in the OWL language, we assume incomplete knowledge of instances.  Any negation of a fact must be explicitly stated.

\textbf{Example 3 (Assumption of ``open world'').}
Consider an example illustrating how making a given assumption about the ``closedness'' or ``openness'' of the world affects inference.
The Table~\ref{tab:gluten} represents information from the knowledge base under development regarding food products and the allergens that may be present in their composition.
\begin{table}[!htb]
    \caption{Food products and the allergens typically found in them.}
    \centering
    \begin{tabular}{|l | l |}
    \hline
    \textbf{Product} & \textbf{Allergens} \\
\hline
    wheat flour & gluten  \\
    dark soy sauce & soya  \\
    sausages &  soya, gluten \\
    cream &  milk \\
    peanuts  & nuts \\
\hline
    \end{tabular}
    \label{tab:gluten}
\end{table}

Let us assume that we want to query the knowledge base on gluten-free products.
Assuming a ``closed world'', the answers would be: \texttt{soy sauce}, \texttt{cream} and \texttt{peanuts}. However, in reality, both cream and soy sauce may contain gluten in their composition, only this information may not have been included in the knowledge base yet. In order to be sure of the results of inference under the assumption of an ``open world'', it would be necessary to include axioms explicitly stating that, for example, peanuts do not contain gluten into the knowledge base.
$\blacksquare$

\subsubsection{Lack of unique names assumption}
Unlike most knowledge representation languages, OWL inference does not use the \emph{Unique Names Assumption, UNA}, which is the assumption that distinct names denote distinct objects.
In order to model knowledge in decentralised environments such as the Internet, it has been assumed that anyone can call anything by any name. Therefore, names cannot be assumed to be unique. However, different names do not necessarily mean different objects either.
For example, two different names, \texttt{Madame Curie} and \texttt{Maria Sklodowska-Curie}, can refer to the same person. In addition, a person may be represented in some knowledge bases by alphanumeric identifiers, e.g.~\texttt{wikidata:Q7186}. 

\textbf{Example 4 (Lack of assumption on uniqueness of names).}
The lack of assumption about uniqueness of names also affects the results of inference using functional properties.
Suppose we have the following axioms in the knowledge base:
\begin{verbatim}
Ola has_father Jan
Ola has_father Marcin
has_father rdf:type owl:FunctionalProperty .
\end{verbatim}

What conclusions will be derived from such a knowledge base? Will the inference engine show a contradiction?
Well, due to the fact that \texttt{has\_father} is the functional property and the lack of assumption of uniqueness of names, when the inference engine is run, a fact will be generated regarding the identity of the instances of \texttt{Jan} and \texttt{Marcin} i.e.:~\texttt{Jan owl:sameAs Marcin}. 
$\blacksquare$

\section{Knowledge graphs}

A \emph{knowledge graph} is a large, graphically structured knowledge base that represents facts in the form of relationships between entities. The basic building blocks of a knowledge graph are: entities, expressed through nodes in the graph, their properties (attributes) and the relationships connecting the nodes, expressed through edges in the graph.
Entities may have (semantic) types, which is represented by the relation \texttt{is-a} between an entity and its type. It is also possible that some types, properties and relationships stored in the knowledge graph are structured in an ontology or data schema.

\begin{figure}[!b]
    \centering
    \includegraphics[width=1.0\textwidth]{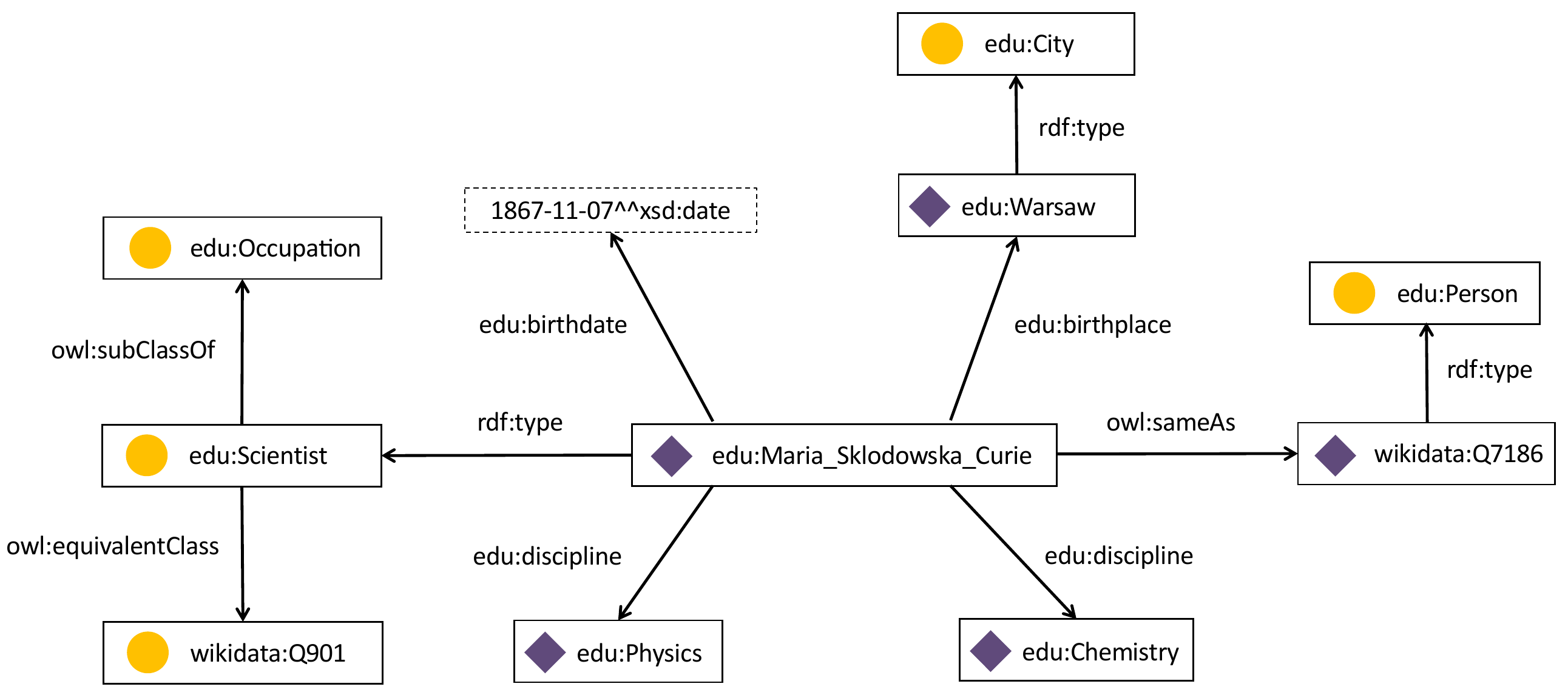}
    \caption{Example of a knowledge graph that contains classes, instances and data of a specific type.\label{fig:kgexample}}.
\end{figure}

\textbf{Example 5 (Knowledge graph).}
Figure~\ref{fig:kgexample} shows an example of a knowledge graph.
The graph uses vocabulary from the RDF, RDFS and OWL namespaces, which are denoted in the figure by the prefixes \texttt{rdf:}, \texttt{rdfs:} and \texttt{owl:} respectively.
The instance \texttt{Maria\_Sklodowska\_Curie} is linked to its semantically equivalent instance \texttt{wikidata:Q7186} (which has an alpha-numeric identifier) in the Wikidata knowledge base via the property \texttt{owl:sameAs}. 
The instance \texttt{wikidata:Q7186} belongs to the class \texttt{Person} and its type (class) is specified by using the property \texttt{rdf:type}. 
The class \texttt{Scientist} is related to the class \texttt{Occupation} through the property \texttt{rdfs:subClassOf}, meaning that it is a subclass of it, as well as through the property \texttt{owl:equivalentClass} to the class \texttt{wikidata:Q901}, meaning that they are semantically equivalent classes. 
In the figure we have examples of both object properties and datatype properties. 
An example of the first of these is the \texttt{discipline} property, which links two objects, among others: \texttt{Maria\_Sklodowska\_Curie} and \texttt{Chemistry}. 
An example of the second of these is the \texttt{birthdate}, which associates the \texttt{Maria\_Sklodowska\_Curie} object with a \texttt{1867-11-07} value that belongs to a specific data type, denoted via the \texttt{xsd:date} namespace as date.
$\blacksquare$

More formal definitions specify a set of entities $\mathcal{E}$, a set of relations $\mathcal{R}$, and a knowledge graph as a directed multi-relational graph $\mathcal{G}$, representing facts or assertions as triples $(s, p, o)$, consisting of a subject $s$, a predicate $p$, and an object $o$, where
\begin{equation}
\mathcal{G} \subseteq \mathcal{E} \times \mathcal{R} \times \mathcal{E},
\end{equation}
\begin{equation}
(s, p, o) \subseteq \mathcal{G}
\end{equation}

We can see a similarity with the already presented examples of semantic networks, the definition of RDF triples, or ontologies.
How, then, does a knowledge graph differ from a graph composed of RDF triples or from an ontology?
Not every knowledge graph uses RDF vocabulary and Semantic Web technologies, although many knowledge graphs are represented using these technologies.
In a knowledge graph, an ontology is a kind of data schema that imposes semantics, meaning on the data, and usually has a shallow level of axiomatisation or is a small part of the knowledge graph. Knowledge graphs, on the other hand, are focused on data (instances) and the number of instances in a typical knowledge graph can be huge. It can be said that a knowledge graph is a kind of semantic network with added constraints.
Due to this much larger scale, also the methods of knowledge acquisition or inference using the knowledge graph have to be adapted to this larger scale. One can observe a shift of attention from manual knowledge engineering methods, focusing on rule modelling and ontologies, to automatic or semi-automatic methods, often using data mining or machine learning or \emph{crowdsourcing}\footnote{Crowdsourcing refers to the practice of obtaining services or content by soliciting input from a large group of people, usually via the Internet.
Crowdsourcing projects often involve breaking larger projects into micro-tasks, separate units of work that can be done independently and quickly, and outsourcing them to a range of collaborators rather than to a single person or organisation. The term is a combination of ``crowd'' and ``outsourcing'' and was coined in 2006 by Jeff Howe
}. Also, knowledge graph inference makes more use of graph data structure and a triple data model than complex logical inference and logical interpretation of data in the form of ontology axioms, and is often performed using statistical or neuro-symbolic methods and learning sub-symbolic representations of knowledge graphs (knowledge graph embeddings~\cite{wang2017knowledge}).

Depending on the availability of the knowledge graph, how it is built (within an organisation or through a community) the result can be an open or corporate knowledge graph. Open knowledge graphs are publicly available. Well known examples of such graphs are DBpedia~\cite{DBLP:conf/semweb/AuerBKLCI07}, Wikidata~\cite{10.1145/3543873.3585579}, YAGO~\cite{DBLP:conf/esws/TanonWS20}.
They cover many domains and offer multilingual lexicalisation. Open knowledge graphs can also address specific domains such as media, geography and others.

Corporate knowledge graphs are internal within a particular company and are aimed at commercial applications. Corporate knowledge graphs are used in industries such as web search, e-commerce, social networks, pharmaceuticals, finance and others.
Typical applications of knowledge graphs include semantic search, question answering, intelligent assistants, innovation support in research and design (such as the design of new drugs).

\subsection{Knowledge graph construction}
\label{sec:kgec}
\begin{figure}[!htb]
    \centering
    \includegraphics[width=0.9\textwidth]{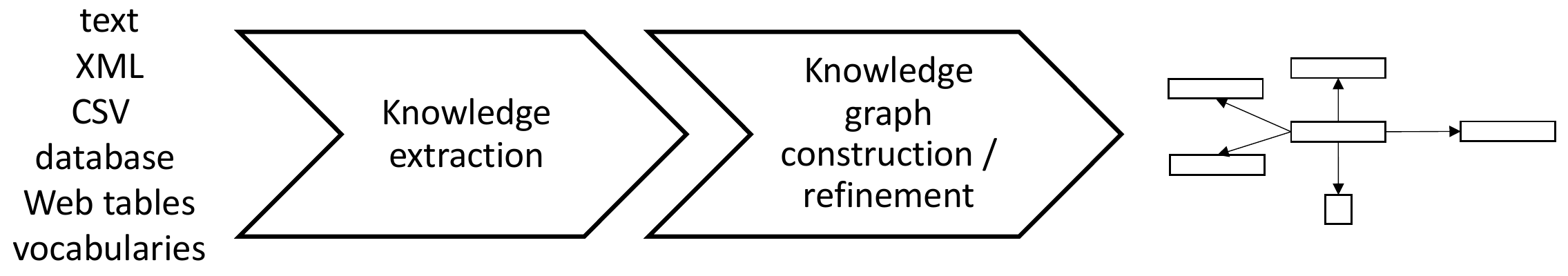}
    \caption{Knowledge graph construction process.}
    \label{fig:construction}
\end{figure}

The typical knowledge graph construction process starts with the acquisition of a corpus and ends with a graph ready for application.
Typically, this process can be illustrated by two main phases: \emph{knowledge extraction} and the construction of the knowledge graph (including its completion or refinement) as illustrated in ~Figure~\ref{fig:construction}.

The first stage involves extracting knowledge from both unstructured (e.g. text) and structured (e.g. relational databases) and semi-structured sources or by converting existing data (e.g. csv files). Natural language processing and information extraction methods can be used for this purpose.
The task of knowledge extraction includes using existing knowledge resources (knowledge bases, ontologies) or generating a schema from source data.

The specific task we may therefore have to deal with at this stage is \emph{entity linking}.
Entity linking is the process of identifying and tagging mentions of specific objects in the text and linking them to the corresponding entities (their identifiers) in external knowledge bases, databases, or ontologies.
For example, in the text \texttt{Jan Kowalski has seen Chicago}, an entity linking system could tag the entities \texttt{Jan Kowalski} and \texttt{Chicago} and link them to the corresponding pages in Wikipedia or identifiers in Wikidata. Sometimes there can be ambiguity as to which specific entity a mention should be linked to. The entity linking system has to deal with the problem of determining whether the text \texttt{Chicago} refers to a city, a musical or a movie. Similarly, it may not be certain whether the mention of ``Jan Kowalski'' refers to an athlete, a writer, etc. In such situations, the system must take this ambiguity into account and help select the appropriate entity to link to the mention in the text.

Once created using knowledge extraction methods, the knowledge graph may contain a lot of noisy and incomplete data.
The purpose of the \emph{knowledge graph completion} task is precisely to fill in missing information and intelligently clean the data in the knowledge graph. This is usually solved by completing missing edges through link prediction, entity deduplication (eliminating repeated entities) and dealing with missing values.

\emph{Link prediction} in knowledge graphs is the problem of predicting missing relationships between entities in a knowledge graph. Missing relationships can be useful to complete the linking of missing information in a knowledge graph or to improve the accuracy of various applications, such as recommender systems or question answering.

There are many different approaches to solving the problem of predicting relationships in knowledge graphs, including regression-based, classification-based and ranking-based approaches. All of these approaches require learning a model on a large dataset containing existing relationships in the knowledge graph, and then using this model to predict missing relationships.

One of the most commonly used approaches is learning knowledge graph embeddings. In this approach, entities and relationships in a knowledge graph are represented as vectors in a low-dimensional vector space, and the model learns these vectors from a large dataset containing existing relationships in the knowledge graph. Then, to predict a missing relationship between two entities, the model compares the vectors of these entities and predicts how probable is that they are connected by a given relationship.

\subsection{Knowledge graph representation learning}

Representation learning involves embedding a data element, which can be, for example, a piece of text, an entity, a relation in a vector space.
\emph{Knowledge graph embedding} is performed using supervised machine learning on a large dataset of triples to project knowledge graph components onto a continuous and low-dimensional vector space.
The aim of knowledge graph embedding is to capture the semantics of entities and relationships in the knowledge graph in a way that will facilitate use in a variety of tasks, be it tasks related to construction of the knowledge graph (such as its completion) or downstream tasks such as its use in recommender systems.

In particular, for any pair ${s,o}\subseteq \mathcal{E}$ and relation $p \in \mathcal{R}$, it can be determined whether the sentence $(s,p,o)$ is true according to the data embeddings of the knowledge graph.

The knowledge graph embedding model consists of:
\begin{itemize}
\item a knowledge graph $\mathcal{G}$,
\item a strategy for generating negative examples,
\item an evaluation function of the triple $f(t)$,
\item a loss function $\mathcal{L}$,
\item a lookup layer,
\item an optimization algorithm.
\end{itemize}

\begin{figure}[!htb]
    \centering
    \includegraphics[width=1.0\textwidth]{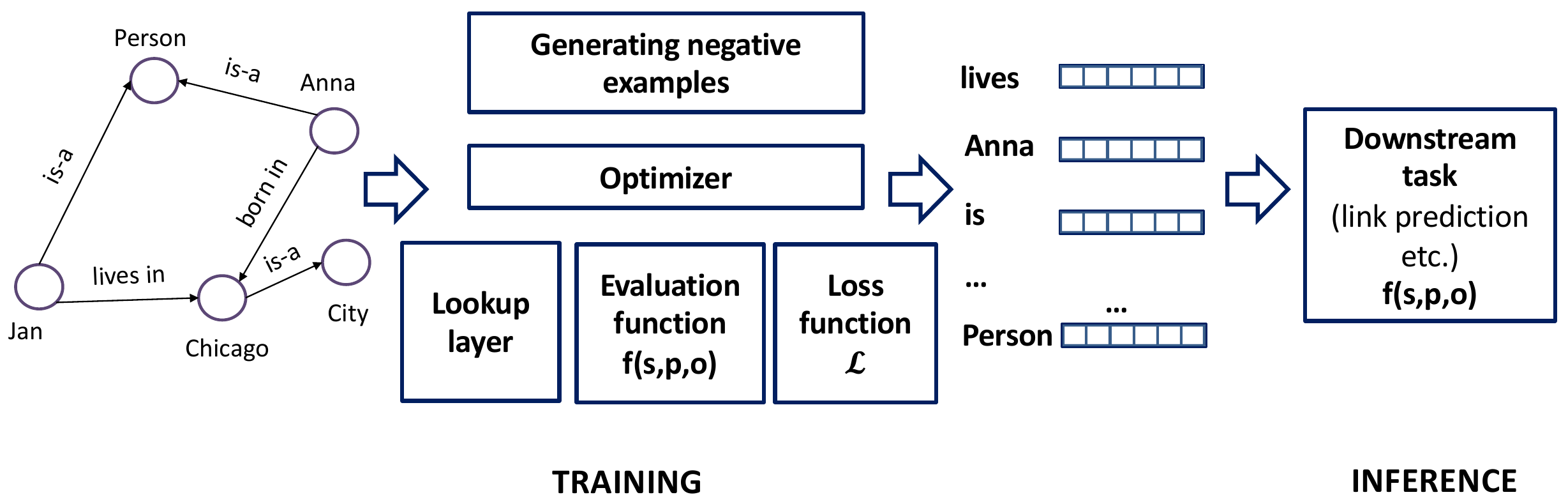}
    \caption{Architecture of a system for creating knowledge graph embeddings.}
    \label{fig:embarch}
\end{figure}

The architecture of such a solution, depicted by the authors of one of the popular libraries for learning knowledge graph representations~\cite{ampligraph}, is shown in Figure~\ref{fig:embarch}.

A number of different approaches to training knowledge graph embeddings have been proposed, including translation-based approaches, tensor-based approaches and graph-based approaches.
For example, the TransE model, one of the first translation-based embedding methods to have been proposed, works on the simple principle that the combination of subject and relation vectors should ideally be equal to the object vector.
TransE is able to learn composition, inverse and antisymmetry.

\section{Large language models as knowledge bases}
As we described in Section~\ref{sec:kgec}, creating and populating knowledge bases requires knowledge extraction from different data formats.
Traditionally, this requires a fixed data schema and the application of NLP methods in several steps, where each step can propagate errors to the next step. 
Petroni et al.~\cite{dblp:conf/emnlp/PetroniRRLBWM19} have shown that instead of the classical approach, \emph{large language models (LLMs)} can be used as a source of knowledge.
Such models, sometimes called \emph{Foundation Models}, are deep neural networks scaled to billions of parameters on the task of predicting the next word on a large corpus and storing the knowledge that was contained in the training data implicitly.  
They can generalise to new tasks without fine-tuning to answer questions structured as ``fill-in-the-blank'' cloze statements, such as: 
``\texttt{The colour of a carrot is [MASK]}''. 

This approach does not require manual engineering of the knowledge schema or human annotation of the data to extract relatively good-quality knowledge.  
Narayan et al.~\cite{DBLP:journals/pvldb/NarayanCOR22} have shown how knowledge cleaning and integration tasks, including entity matching, can be performed by reformulating them as prompting tasks. For example, they examined the answer to the question ``Are products A and B the same?'' and the language model generated a string ``Yes'' or ``No'' as the answer. 

\subsection{Prompt engineering}
\emph{Prompt engineering} (\emph{in-context prompting})~\cite{DBLP:conf/acl/BeltagyCLMC22} concerns methods of communicating with LLMs to get desired answers. The weights of the model stay unchanged. 
The models can be trained to learn a task in a few-shot manner (with minimal task description) or even in a zero-shot learning setting~\cite{DBLP:journals/corr/abs-2005-14165}. 
Prompt engineering is mostly an experimental field of study. The terminology is relatively simple and is introduced next. 
\emph{Prompt} is a conditioning text before the test input.
Zero-shot learning consists of simply providing a text to get the answer.
\emph{Demonstrations} is an instance of the prompt, which is a concatenation of the $k$-shot training
data. 
Few-shot learning consists of presenting a set of demonstrations composed of input and envisaged output, for instance: 
\begin{verbatim}
Lemon: yellow
Carrot: orange
Raspberry: raspberry red
Pear: 
\end{verbatim}
\emph{Pattern} is a function mapping an input to the text. 
\emph{Verbalizer} is a function mapping a label to the text. 

\subsection{Commonsense knowledge}
Some datasets and knowledge graphs are designed to capture commonsense knowledge, such as the knowledge of physical phenomena that humans acquire during their lives as part of their interaction with the environment or knowledge required in scientific computing.  
For example, ConceptNet~\cite{dblp:conf/aaai/SpeerCH17} is a knowledge graph that connects structured knowledge to natural language, bridging the gap between formal knowledge representation and natural language. 
It connects words and phrases with labelled edges. It gathers knowledge from the resources developed by experts, crowd-sourcing, and games with a purpose. 
Further linking this graph with text embeddings helps to solve tasks such as SAT-style analogies more efficiently than with resources based primarily on formalised knowledge structure. 

Benchmarks designed to evaluate commonsense reasoning often come with the task of question answering. For instance, PIQA~(``Physical Interaction -- Question Answering'')~\cite{bisk2020piqa} is a dataset related to best achieving a goal in everyday tasks such as crafting, baking, or manipulating objects using everyday materials. The user has to choose one of the two answers concerning how to achieve the goal best. One of the answers is the right one. For instance, if one asks how to eat soup, the correct answer is to use a spoon rather than a fork. 
Answering such questions requires knowledge that may not be represented explicitly as, e.g., attributes of some object (as factual knowledge) and also might depend on context.  
Consider, for instance, querying an LLM on the functions (roles) of some ingredients in a dish (implicit and contextual knowledge): 
\begin{verbatim}
   sugar [DISH] cake [FUNCTION] sweetener
   baking powder [DISH] cake [FUNCTION] leavening agent
   egg yolk [DISH] Hollandaise sauce [FUNCTION] emulsifier
   yeast [DISH] bread [FUNCTION]  
\end{verbatim}
Answering such queries may require the model to understand food technological processes. 

\section{Knowledge engineering methodology}
There are a number of methodological issues and best practices associated with knowledge engineering. Several of these are discussed below.

\subsection{URIs and multilinguality}
We can adopt one of two main naming conventions when specifying a URI for a resource. 

One is to name the resource directly in its URI, i.e.,~insert the resource name in the URI string, such as~\texttt{geo:Warsaw}.
The advantages of such a solution are simplicity, ease of interpretation and greater availability of tools that support such a format without problems.

The second convention is to create an alphanumeric resource identifier and put the name in the label using predefined annotation properties like~\texttt{rdfs:label}, for example: \texttt{geo:locality1 rdfs:label "Warsaw"}.
The advantages of such a solution are: facilitating multilingualism, and avoiding the problems of formulating long names, which can be, for example, a concatenation of many words due to the addition of more and more detailed classes in their hierarchy (we can put any phrase in the label), ease and flexibility of changes, and preserving the stability of URI-based application builds in case the semantic meaning of a resource changes or drifts (it is then enough to change the label itself and not the identifier).

\subsection{Ontology engineering methodologies}
Knowledge engineering refers to the process of developing knowledge bases, ontologies and knowledge graphs.
In particular, ontology engineering methodologies have been developed.
Early methodologies followed the cascade model of ontology development, in which requirements and general conceptualization were established before ontology definition began.  However, nowadays we often deal with large or constantly evolving ontologies, hence the subsequent methodologies that have been proposed promote more iterative and agile ways of building and maintaining ontologies~\cite{DBLP:conf/esws/KeetL16}.

Modern methodologies propose various best practices like reusing already existing resources to create a network of ontologies that import selected classes, properties or entire modules among themselves~\cite{DBLP:books/daglib/p/Suarez-FigueroaGMG12}.
They often contain two common elements: ontology requirements and ontology design patterns~\cite{gangemi2005ontology}.

A common way of expressing the requirements that an ontology must meet includes \emph{competency questions}, which are natural language questions that an ontology or knowledge graph should be prepared to answer, i.e., at least contain the appropriate vocabulary.

\emph{Ontology Design Patterns} define general ontology modelling patterns (modelling templates) that can be used as inspiration for modelling more specific phenomena. An example would be a pattern defining arbitrary events, which models, among other things, the spatial and temporal scope and the participants of the event and is supplemented with competency questions and other documentation elements.

\subsection{FAIR principles}
\emph{FAIR Principles} were originally proposed in the context of publishing scientific data~\cite{wilkinson2016fair}, but are generally applicable to any situation where data is to be open, accessible, and published in a way that facilitates its reuse by external parties, with a particular emphasis on facilitating its processing in information systems.
The FAIR acronym refers to four fundamental principles for data, metadata or both, each with specific purposes.
Below is a brief discussion of each of the FAIR principles:
\begin{itemize}
\item \emph{findable}: research results should be discoverable and easy to locate, using persistent identifiers (e.g., DOIs) and metadata that accurately and comprehensively describe the content.
\item \emph{accessible}: research results should be available to anyone who wants to access them, regardless of location or ability to pay. This can be achieved through open-access publishing or other mechanisms that provide free and unrestricted access.
\item \emph{interoperable}: research results should be structured and formatted in a way that allows them to be easily integrated and linked with other data and research results. This requires the use of common standards and protocols.
\item \emph{reusable}: research results should be licensed in a way that allows them to be reused and built upon by others, subject to proper attribution and citation.
\end{itemize}
Adhering to FAIR can help make scientific research more transparent, reproducible, and influential, advancing science and benefiting society.

\section{Further reading}
A more detailed description of the RDF model can be found in Heath and Bizer's book ``Linked Data: Evolving the Web into a Global Data Space''~\cite{Heath2011}.

An introduction to description logic can be found in the article by Kr{o}tzsch et al.~\cite{DBLP:journals/expert/KrotzschSH14}.
A comprehensive textbook on description logic is by Baader et al.~\cite{DBLP:conf/dlog/2003handbook}.
Description logics are also discussed in a book by Hitzler et al. ``Foundations of Semantic Web Technologies''~\cite{books/crc/Hitzler2010}, on the theoretical foundations of the technologies.

Ontology modelling including topics of engineering, good practices, and design patterns is presented in the books: ``Semantic Web for the Working Ontologist: Effective Modeling for Linked Data, RDFS, and OWL'' by Allemang, Hendler and Gandon~\cite{10.1145/3382097} and ``Demystifying OWL for the Enterprise'' by Uschold~\cite{DBLP:series/synthesis/2018Uschold}. Knowledge engineering  (theoretical foundations of ontology representation languages and good modelling practices) is covered in the publicly available textbook ``An Introduction to Ontology Engineering'' by Keet~\cite{mariabook2018}.

Knowledge graphs constitute a relatively new, active, and interdisciplinary area of artificial intelligence that emerged around 2012. They draw from areas such as natural language processing, data mining, and the Semantic Web.
There are relatively recent textbooks, including ``Knowledge Graphs: Fundamentals, Techniques, and Applications'' by Kejriwal, Knoblock and Szekely~\cite{kejriwal2020} and ``Knowledge Graphs'' by Hogan et al.~\cite{kg-book}.
\label{sec:bibliography}
\\\\
\small
\textbf{Acknowledgments}. 
I thank Robert Nowak, Mieczysław Muraszkiewicz and Students for helpful remarks on previous versions of this text. 

\bibliographystyle{plain} 

\begin{thebibliography}{10}

\bibitem{10.1145/3382097}
Dean Allemang, James Hendler, and Fabien Gandon.
\newblock {\em Semantic Web for the Working Ontologist: Effective Modeling for
  Linked Data, RDFS, and OWL}.
\newblock Association for Computing Machinery, 2020.

\bibitem{10.5555/2846229}
Robert Arp, Barry Smith, and Andrew~D. Spear.
\newblock {\em Building Ontologies with Basic Formal Ontology}.
\newblock The MIT Press, 2015.

\bibitem{DBLP:conf/semweb/AuerBKLCI07}
S{\"{o}}ren Auer, Christian Bizer, Georgi Kobilarov, Jens Lehmann, Richard
  Cyganiak, and Zachary~G. Ives.
\newblock {DB}pedia: {A} nucleus for a web of open data.
\newblock In Karl Aberer, Key{-}Sun Choi, Natasha~Fridman Noy, Dean Allemang,
  Kyung{-}Il Lee, Lyndon J.~B. Nixon, Jennifer Golbeck, Peter Mika, Diana
  Maynard, Riichiro Mizoguchi, Guus Schreiber, and Philippe
  Cudr{\'{e}}{-}Mauroux, editors, {\em The Semantic Web, 6th International
  Semantic Web Conference, 2nd Asian Semantic Web Conference, {ISWC} 2007 +
  {ASWC} 2007, Busan, Korea, November 11-15, 2007}, volume 4825 of {\em Lecture
  Notes in Computer Science}, pages 722--735. Springer, 2007.

\bibitem{DBLP:conf/dlog/2003handbook}
Franz Baader, Diego Calvanese, Deborah~L. McGuinness, Daniele Nardi, and
  Peter~F. Patel{-}Schneider, editors.
\newblock {\em The Description Logic Handbook: Theory, Implementation, and
  Applications}.
\newblock Cambridge University Press, 2003.

\bibitem{10.3115/980845.980860}
Collin~F. Baker, Charles~J. Fillmore, and John~B. Lowe.
\newblock The {B}erkeley {F}rame{N}et project.
\newblock In {\em Proceedings of the 36th Annual Meeting of the Association for
  Computational Linguistics and 17th International Conference on Computational
  Linguistics - Volume 1}, ACL '98/COLING '98, page 86–90, USA, 1998.
  Association for Computational Linguistics.

\bibitem{DBLP:conf/acl/BeltagyCLMC22}
Iz~Beltagy, Arman Cohan, Robert L.~Logan IV, Sewon Min, and Sameer Singh.
\newblock Zero- and few-shot {NLP} with pretrained language models.
\newblock In Luciana Benotti, Naoaki Okazaki, Yves Scherrer, and Marcos
  Zampieri, editors, {\em Proceedings of the 60th Annual Meeting of the
  Association for Computational Linguistics, {ACL} 2022 - Tutorial Abstracts,
  Dublin, Ireland, May 22-27, 2022}, pages 32--37. Association for
  Computational Linguistics, 2022.

\bibitem{berners2001semantic}
Tim Berners-Lee, James Hendler, and Ora Lassila.
\newblock The {S}emantic {W}eb.
\newblock {\em Scientific American}, 284(5):34--43, 2001.

\bibitem{DBLP:series/faia/BesoldGBBDHKLLPPPZ21}
Tarek~R. Besold, Artur~S. d'Avila Garcez, Sebastian Bader, Howard Bowman,
  Pedro~M. Domingos, Pascal Hitzler, Kai{-}Uwe K{\"{u}}hnberger, Lu{\'{\i}}s~C.
  Lamb, Priscila Machado~Vieira Lima, Leo de~Penning, Gadi Pinkas, Hoifung
  Poon, and Gerson Zaverucha.
\newblock Neural-symbolic learning and reasoning: {A} survey and
  interpretation.
\newblock In Pascal Hitzler and Md.~Kamruzzaman Sarker, editors, {\em
  Neuro-Symbolic Artificial Intelligence: The State of the Art}, volume 342 of
  {\em Frontiers in Artificial Intelligence and Applications}, pages 1--51.
  {IOS} Press, 2021.

\bibitem{bisk2020piqa}
Yonatan Bisk, Rowan Zellers, Jianfeng Gao, Yejin Choi, et~al.
\newblock {PIQA}: Reasoning about physical commonsense in natural language.
\newblock In {\em Proceedings of the AAAI conference on artificial
  intelligence}, volume~34, pages 7432--7439, 2020.

\bibitem{kgcookbook}
Andreas Blumauer and Helmut Nagy.
\newblock {\em The Knowledge Graph Cookbook}.
\newblock edition mono/monochrom, 2020.

\bibitem{BrachmanLevesque04}
Ronald Brachman and Hector Levesque.
\newblock {\em Knowledge Representation and Reasoning}.
\newblock The Morgan Kaufmann Series in Artificial Intelligence. Morgan
  Kaufmann, Amsterdam, 2004.

\bibitem{DBLP:journals/corr/abs-2005-14165}
Tom~B. Brown, Benjamin Mann, Nick Ryder, Melanie Subbiah, Jared Kaplan,
  Prafulla Dhariwal, Arvind Neelakantan, Pranav Shyam, Girish Sastry, Amanda
  Askell, Sandhini Agarwal, Ariel Herbert{-}Voss, Gretchen Krueger, Tom
  Henighan, Rewon Child, Aditya Ramesh, Daniel~M. Ziegler, Jeffrey Wu, Clemens
  Winter, Christopher Hesse, Mark Chen, Eric Sigler, Mateusz Litwin, Scott
  Gray, Benjamin Chess, Jack Clark, Christopher Berner, Sam McCandlish, Alec
  Radford, Ilya Sutskever, and Dario Amodei.
\newblock Language models are few-shot learners.
\newblock {\em CoRR}, abs/2005.14165, 2020.

\bibitem{ampligraph}
Luca Costabello, Sumit Pai, Chan~Le Van, Rory McGrath, Nick McCarthy, and Pedro
  Tabacof.
\newblock {AmpliGraph: a Library for Representation Learning on Knowledge
  Graphs}, 2019.

\bibitem{davis1977production}
Randall Davis, Bruce Buchanan, and Edward Shortliffe.
\newblock Production rules as a representation for a knowledge-based
  consultation program.
\newblock {\em Artificial intelligence}, 8(1):15--45, 1977.

\bibitem{gangemi2005ontology}
Aldo Gangemi.
\newblock Ontology design patterns for semantic web content.
\newblock In {\em The Semantic Web--ISWC 2005: 4th International Semantic Web
  Conference, ISWC 2005, Galway, Ireland, November 6-10, 2005. Proceedings 4},
  pages 262--276. Springer, 2005.

\bibitem{gangemi2002dolce}
Aldo Gangemi, Nicola Guarino, Claudio Masolo, Alessandro Oltramari, and Luc
  Schneider.
\newblock {\em Sweetening Ontologies with DOLCE}, pages 166--181.
\newblock Springer, Berlin, Heidelberg, 2002.

\bibitem{Gruber1993ATA}
Thomas Gruber.
\newblock A translation approach to portable ontology specifications.
\newblock {\em Knowledge Acquisition}, pages 199--220, 1993.

\bibitem{Guarino1998}
Nicola Guarino.
\newblock Formal ontology and information systems.
\newblock In {\em Proceedings of Formal Ontology in Information System}, pages
  3--15. IOS Press, 1998.

\bibitem{Heath2011}
Tom Heath and Christian Bizer.
\newblock {\em Linked Data: Evolving the Web into a Global Data Space}.
\newblock Morgan \& Claypool, 2011.

\bibitem{books/crc/Hitzler2010}
Pascal Hitzler, Markus Krotzsch, and Sebastian Rudolph.
\newblock {\em Foundations of Semantic Web Technologies}.
\newblock Chapman and Hall/CRC Press, 2010.

\bibitem{DBLP:journals/csur/HoganBCdMGKGNNN21}
Aidan Hogan, Eva Blomqvist, Michael Cochez, Claudia d'Amato, Gerard de~Melo,
  Claudio Gutierrez, Sabrina Kirrane, Jos{\'{e}} Emilio~Labra Gayo, Roberto
  Navigli, Sebastian Neumaier, Axel{-}Cyrille~Ngonga Ngomo, Axel Polleres,
  Sabbir~M. Rashid, Anisa Rula, Lukas Schmelzeisen, Juan~F. Sequeda, Steffen
  Staab, and Antoine Zimmermann.
\newblock Knowledge graphs.
\newblock {\em {ACM} Comput. Surv.}, 54(4):71:1--71:37, 2022.

\bibitem{kg-book}
Aidan Hogan, Eva Blomqvist, Michael Cochez, Claudia d'Amato, Gerard de~Melo,
  Claudio Guti\'errez, Sabrina Kirrane, Jos\'e~Emilio Labra~Gayo, Roberto
  Navigli, Sebastian Neumaier, Axel-Cyrille Ngonga~Ngomo, Axel Polleres,
  Sabbir~M. Rashid, Anisa Rula, Lukas Schmelzeisen, Juan~F. Sequeda, Steffen
  Staab, and Antoine Zimmermann.
\newblock {\em {K}nowledge {G}raphs}.
\newblock Number~22 in Synthesis Lectures on Data, Semantics, and Knowledge.
  Springer, 2021.

\bibitem{mariabook2018}
C.~Maria Keet.
\newblock {\em An Introduction to Ontology Engineering}.
\newblock College Publications, 2018.

\bibitem{DBLP:conf/esws/KeetL16}
C.~Maria Keet and Agnieszka Lawrynowicz.
\newblock Test-driven development of ontologies.
\newblock In Harald Sack, Eva Blomqvist, Mathieu d'Aquin, Chiara Ghidini,
  Simone~Paolo Ponzetto, and Christoph Lange, editors, {\em The Semantic Web.
  Latest Advances and New Domains - 13th International Conference, {ESWC} 2016,
  Heraklion, Crete, Greece, May 29 - June 2, 2016, Proceedings}, volume 9678 of
  {\em Lecture Notes in Computer Science}, pages 642--657. Springer, 2016.

\bibitem{kejriwal2020}
Mayank Kejriwal, Craig Knoblock, and Pedro Szekely.
\newblock {\em Knowledge Graphs: Fundamentals, Techniques, and Applications}.
\newblock The MIT Press, 2021.

\bibitem{inbookK1}
Robert Kowalski.
\newblock {\em Logic for Problem Solving, Revisited.}
\newblock Books OnDemand Gmbh, 2014.

\bibitem{inbookK2}
Robert Kowalski.
\newblock {\em Computational Logic and Human Thinking: How to Be Artificially Intelligent.}
\newblock Cambridge University Press, 2010.

\bibitem{DBLP:journals/expert/KrotzschSH14}
Markus Krötzsch, František Simančík, and Ian Horrocks.
\newblock Description Logics.
\newblock {\em {IEEE} Intelligent Systems}, pages 12--19, 2014.

\bibitem{10.5555/889222}
Marvin Minsky.
\newblock A framework for representing knowledge.
\newblock Technical report, USA, 1974.

\bibitem{DBLP:journals/pvldb/NarayanCOR22}
Avanika Narayan, Ines Chami, Laurel~J. Orr, and Christopher R{\'{e}}.
\newblock Can foundation models wrangle your data?
\newblock {\em Proc. {VLDB} Endow.}, 16(4):738--746, 2022.

\bibitem{newell1956logic}
Allen Newell and Herbert Simon.
\newblock The logic theory machine--a complex information processing system.
\newblock {\em IRE Transactions on information theory}, 2(3):61--79, 1956.

\bibitem{NonakaTakeuchi1995aa}
Ikujirō Nonaka and Hirotaka Takeuchi.
\newblock {\em The Knowledge-Creating Company: How Japanese Companies Create
  the Dynamics of Innovation}.
\newblock Oxford University Press, 1995.

\bibitem{dblp:conf/emnlp/PetroniRRLBWM19}
Fabio Petroni, Tim Rockt{\"{a}}schel, Sebastian Riedel, Patrick S.~H. Lewis,
  Anton Bakhtin, Yuxiang Wu, and Alexander~H. Miller.
\newblock Language models as knowledge bases?
\newblock In Kentaro Inui, Jing Jiang, Vincent Ng, and Xiaojun Wan, editors,
  {\em Proceedings of the 2019 Conference on Empirical Methods in Natural
  Language Processing and the 9th International Joint Conference on Natural
  Language Processing, {EMNLP-IJCNLP} 2019, Hong Kong, China, November 3-7,
  2019}, pages 2463--2473. Association for Computational Linguistics, 2019.

\bibitem{richens1956preprogramming}
Richard~H Richens.
\newblock Preprogramming for mechanical translation.
\newblock {\em Mech. Transl. Comput. Linguistics}, 3(1):20--25, 1956.

\bibitem{dblp:conf/aaai/SpeerCH17}
Robyn Speer, Joshua Chin, and Catherine Havasi.
\newblock Conceptnet 5.5: An open multilingual graph of general knowledge.
\newblock In Satinder Singh and Shaul Markovitch, editors, {\em Proceedings of
  the Thirty-First {AAAI} Conference on Artificial Intelligence, February 4-9,
  2017, San Francisco, California, {USA}}, pages 4444--4451. {AAAI} Press,
  2017.

\bibitem{DBLP:books/daglib/p/Suarez-FigueroaGMG12}
Mari~Carmen Su{\'{a}}rez{-}Figueroa, Asunci{\'{o}}n G{\'{o}}mez{-}P{\'{e}}rez,
  Enrico Motta, and Aldo Gangemi.
\newblock Introduction: Ontology engineering in a networked world.
\newblock In Mari~Carmen Su{\'{a}}rez{-}Figueroa, Asunci{\'{o}}n
  G{\'{o}}mez{-}P{\'{e}}rez, Enrico Motta, and Aldo Gangemi, editors, {\em
  Ontology Engineering in a Networked World}, pages 1--6. Springer, 2012.

\bibitem{DBLP:conf/esws/TanonWS20}
Thomas~Pellissier Tanon, Gerhard Weikum, and Fabian~M. Suchanek.
\newblock {YAGO} 4: {A} reason-able knowledge base.
\newblock In Andreas Harth, Sabrina Kirrane, Axel{-}Cyrille~Ngonga Ngomo, Heiko
  Paulheim, Anisa Rula, Anna~Lisa Gentile, Peter Haase, and Michael Cochez,
  editors, {\em The Semantic Web - 17th International Conference, {ESWC} 2020,
  Heraklion, Crete, Greece, May 31-June 4, 2020, Proceedings}, volume 12123 of
  {\em Lecture Notes in Computer Science}, pages 583--596. Springer, 2020.

\bibitem{tarski1954}
Alfred Tarski.
\newblock Contributions to the theory of models. I. 
\newblock {\em Indagationes Mathematicae}, 16:572–581, 1954.

\bibitem{DBLP:series/synthesis/2018Uschold}
Michael Uschold.
\newblock {\em Demystifying {OWL} for the Enterprise}.
\newblock Morgan {\&} Claypool Publishers, 2018.

\bibitem{Uschold1996}
Michael Uschold and Michael Gruninger.
\newblock Ontologies: principles, methods and applications.
\newblock {\em The Knowledge Engineering Review}, pages 93--136, 1996.

\bibitem{10.1145/3543873.3585579}
Denny Vrande\v{c}i\'{c}, Lydia Pintscher, and Markus Kr\"{o}tzsch.
\newblock Wikidata: The making of.
\newblock In {\em Companion Proceedings of the ACM Web Conference 2023}, WWW'23
  Companion, page 615–624, New York, NY, USA, 2023. Association for Computing
  Machinery.

\bibitem{wang2017knowledge}
Quan Wang, Zhendong Mao, Bin Wang, and Li~Guo.
\newblock Knowledge graph embedding: A survey of approaches and applications.
\newblock {\em IEEE Transactions on Knowledge and Data Engineering},
  29(12):2724--2743, 2017.

\bibitem{wilkinson2016fair}
Mark~D. Wilkinson., Michel Dumontier, IJsbrand~Jan Aalbersberg, Gabrielle
  Appleton, Myles Axton, Arie Baak, Niklas Blomberg, Jan-Willem Boiten,
  Luiz~Bonino da~Silva~Santos, Philip~E. Bourne, Jildau Bouwman, Anthony~J.
  Brookes, Tim Clark, Merc\`e Crosas, Ingrid Dillo, Olivier Dumon, Scott
  Edmunds, Chris~T. Evelo, Richard Finkers, Alejandra Gonzalez-Beltran,
  Alasdair~J.G. Gray, Paul Groth, Carole Goble, Jeffrey~S. Grethe, Jaap
  Heringa, Peter~A.C ’t Hoen, Rob Hooft, Tobias Kuhn, Ruben Kok, Joost Kok,
  Scott~J. Lusher, Maryann~E. Martone, Albert Mons, Abel~L. Packer, Bengt
  Persson, Philippe Rocca-Serra, Marco Roos, Rene van Schaik, Susanna-Assunta
  Sansone, Erik Schultes, Thierry Sengstag, Ted Slater, George Strawn,
  Morris~A. Swertz, Mark Thompson, Johan van~der Lei, Erik van Mulligen, Jan
  Velterop, Andra Waagmeester, Peter Wittenburg, Katherine Wolstencroft, Jun
  Zhao, and Barend Mons.
\newblock {The FAIR Guiding Principles for scientific data management and
  stewardship}.
\newblock {\em Scientific data}, pages 1--9, 2016.


\end{thebibliography}

\end{document}